\documentclass[preprint,12pt,authoryear]{elsarticle}

\usepackage{amssymb}
\usepackage{url}
\usepackage{amsmath}
\usepackage{booktabs}
\usepackage{multirow}
\usepackage{threeparttable}
\usepackage{color}
\usepackage{float}
\usepackage{appendix}
\usepackage {setspace}
\usepackage{algorithm}%
\usepackage{algorithmicx}%
\usepackage{algpseudocode}%

\usepackage{amsthm}

\journal{arXiv}

\begin{document}

\begin{frontmatter}

\title{Reinforcement Learning (RL) Meets Urban Climate Modeling: Investigating the Efficacy and Impacts of RL-Based HVAC Control}

\author[1]{Junjie Yu}

\author[2]{John S. Schreck}

\author[2]{David John Gagne}

\author[3]{Keith W. Oleson}

\author[4]{Jie Li}

\author[5]{Yongtu Liang}

\author[5]{Qi Liao}

\author[6]{Mingfei Sun}

\author[1]{David O. Topping}

\author[1]{Zhonghua Zheng}

\affiliation[1]{organization={Department of Earth and Environmental Sciences, The University of Manchester}, city={Manchester}, postcode={M13 9PL}, country={UK}}

\affiliation[2]{organization={Computational and Information Systems Laboratory, NSF National Center for Atmospheric Research (NCAR)}, city={Boulder}, state={CO}, postcode={80307}, country={USA}}

\affiliation[3]{organization={Climate and Global Dynamics Laboratory, NSF National Center for Atmospheric Research (NCAR)}, city={Boulder}, state={CO}, postcode={80307}, country={USA}}

\affiliation[4]{organization={Centre for Process Integration, Department of Chemical Engineering, The University of Manchester}, city={Manchester}, postcode={M13 9PL}, country={UK}}

\affiliation[5]{organization={Beijing Key Laboratory of Urban Oil and Gas Distribution Technology, China University of Petroleum--Beijing}, city={Beijing}, postcode={102249}, country={China}}

\affiliation[6]{organization={Department of Computer Science, The University of Manchester}, city={Manchester}, postcode={M13 9PL}, country={UK}}

\begin{abstract}

Reinforcement learning (RL)-based heating, ventilation, and air conditioning (HVAC) control has emerged as a promising technology for reducing building energy consumption while maintaining indoor thermal comfort. However, the efficacy of such strategies is influenced by the background climate and their implementation may potentially alter both the indoor climate and local urban climate. This study proposes an integrated framework combining RL with an urban climate model that incorporates a building energy model, aiming to evaluate the efficacy of RL-based HVAC control across different background climates, impacts of RL strategies on indoor climate and local urban climate, and the transferability of RL strategies across cities. Our findings reveal that the reward (defined as a weighted combination of energy consumption and thermal comfort) and the impacts of RL strategies on indoor climate and local urban climate exhibit marked variability across cities with different background climates. The sensitivity of reward weights and the transferability of RL strategies are also strongly influenced by the background climate. Cities in hot climates tend to achieve higher rewards across most reward weight configurations that balance energy consumption and thermal comfort, and those cities with more varying atmospheric temperatures demonstrate greater RL strategy transferability. These findings underscore the importance of thoroughly evaluating RL-based HVAC control strategies in diverse climatic contexts. This study also provides a new insight that city-to-city learning will potentially aid the deployment of RL-based HVAC control. 

\end{abstract}



\end{frontmatter}


\section{Introduction}\label{intro}

Urban development and human activities contribute to the urban heat island (UHI) effect \citep{YANGETAL16ProcediaEngineering, KALNAYCAI03Nature}, which warms urban areas and negatively impacts urban ecology \citep{URBANETAL24NatureClimateChange}, soil temperature and moisture content \citep{SHIETAL12EnvironmentalEarthSciences}, and human health \citep{SINGHETAL20UrbanEcology}. Among the various contributions to the UHI effect, anthropogenic heat has been identified as a key driver \citep{CRUTZEN04Atmos.Environ.}, with building services in urban areas constituting a primary source \citep{SMITHETAL09TheoreticalandAppliedClimatology, LIUETAL22EnergyBuild., GUNERALPETAL17Proc.Natl.Acad.Sci.}. In particular, heating, ventilation, and air conditioning (HVAC) systems---essential for maintaining comfortable indoor environments---constitute a substantial portion of building energy consumption \citep{ENTERIAMIZUTANI11Renew.Sustain.EnergyRev., GUNERALPETAL17Proc.Natl.Acad.Sci.}. In urban commercial buildings, HVAC systems can consume 40\% of the total energy use \citep{edenhofer2015climate, KIMETAL19EnergyBuild.}. 

It is important to note that the structure of building energy demand is expected to shift due to the changing climate \citep{YALEWETAL20NatureEnergy, RAHIFETAL22Build.Environ.}. As global temperatures rise, the use of air conditioning will become more frequent, with a particularly pronounced effect in warmer regions. Conversely, in colder regions, heating demand may decrease. For instance, it is shown that Brussels, the capital of Belgium, will shift from a heating-dominated to a cooling-dominated city forced by climate change \citep{RAHIFETAL22Build.Environ.}. The changing climate will challenge the energy system \citep{DALEETAL15ClimaticChangea} and the changing energy system will, in return, pose risks to the local urban climate due to the extra heat emission, which is a complex dynamic feedback between the building energy system and climate \citep{LIETAL24NatureClimateChange}. This complex dynamic underscores the urgent need to advance HVAC systems to better adapt to a changing climate \citep{WISDOMEBIRIMETAL24Int.j.appl.res.soc.sci.}.
Accordingly, improving HVAC efficiency offers a practical solution, enabling energy savings and thermal comfort without additional infrastructure \citep{PARKKIM24CaseStud.Therm.Eng., VAKILOROAYAETAL14EnergyConvers.Manag., ILDIRIETAL25EnergyandBuildings}, thus providing a feasible and flexible approach for reducing energy consumption and heat emission in urban areas. 

However, developing HVAC control strategies that balance energy efficiency with thermal comfort presents a complex challenge \citep{VAKILOROAYAETAL14EnergyConvers.Manag.}. In recent years, reinforcement learning (RL) has emerged as a promising approach for HVAC control \citep{WEIETAL17Proc.54thAnnu.Des.Autom.Conf.2017, YUETAL21IEEETrans.SmartGrid, RAMANETAL202020Am.ControlConf.ACC, BIEMANNETAL21AppliedEnergy, NGUYENETAL24EnergyBuild., KADAMALAETAL24SmartEnergya, MARKOWITZDRENKOW23ICLR2023WorkshopTacklingClim.ChangeMach.Learn., ZHANGETAL19Proc.6thACMInt.Conf.Syst.Energy-Effic.Build.CitiesTransp.b}, as it can be directly trained to decide when to activate HVAC control. Unlike traditional supervised learning or optimization methods, RL is designed to handle multi-step decision-making and optimize long-term rewards, making it more suitable for practical HVAC control. RL has gained from deep learning \citep{MNIHETAL15Naturea} and the combination of deep learning and reinforcement learning is known as deep reinforcement learning (DRL) \citep{LIETAL23NatureReviewsEarth&Environment}. In DRL, a neural network is used to construct an agent that interacts with the environment to achieve trial and error interaction, from which the agent obtains experience and is optimized toward optimal strategies (maximizing the rewards that the agent obtains from the environment). Once the optimal policy is learned, the agent can be deployed to obtain an optimized sequence policy for HVAC control. 

RL-based HVAC control strategies are typically developed using the building energy model (BEM) as an interactive environment \citep{JIMENEZ-RABOSOETAL21Proc.8thACMInt.Conf.Syst.Energy-Effic.Build.CitiesTransp., LIUETAL24Proc.11thACMInt.Conf.Syst.Energy-Effic.Build.CitiesTransp., MARKOWITZDRENKOW23ICLR2023WorkshopTacklingClim.ChangeMach.Learn.}, which simulates building energy balances and calculates energy consumption. Building energy consumption can be influenced by both indoor climate and urban local climate \citep{LIETAL24J.Adv.Model.EarthSyst.a}. However, the building scale BEM typically focuses on indoor climate and does not account for the two-way interaction between indoor climate and local urban climate. Thus, assessments of the impacts of RL-based HVAC control strategies on local climate remain unexplored. The lack of this interaction will underestimate the risks to urban energy and local urban climates \citep{LIETAL24NatureClimateChange}. 
According to model projections, the omission of two-way feedback results in a 40\% underestimation in summer cooling energy and a 10\% overestimation in winter heating energy \citep{LIETAL24NatureClimateChange}. A model simulation of heat pump (a HAVC technology) also shows that there will be a decrease in near-surface air temperature of up to 0.5 °C after implementing the new HAVC technology \citep{MEYERETAL24NatureCommunications}. 
Therefore, it is essential to quantify the impact of RL-based HVAC control on local urban climate using the urban climate model coupled with the BEM, as the urban climate model coupled with the BEM can provide insight into the impacts of HVAC control systems on local climate. Furthermore, the background climate plays a critical role in shaping the effectiveness of urban climate adaptation strategies. For instance, the performance of interventions such as cool roofs and urban greening has been shown to strongly depend on the prevailing urban climate conditions \citep{WANGETAL20Geophys.Res.Lett.b, ZHAOETAL23Environ.Res.Lett.}.
These collectively highlight the importance of accounting for the complex interactions between indoor climate---affected by the HVAC strategies---and local urban climate, as well as the critical role of the background climate. By coupling RL-based HVAC control with an urban climate model, the urban climate model's inherent capabilities also enable effective exploration of how background climate affects the efficacy and transferability of RL strategies. 

This study aims to establish a framework to quantify the effectiveness of RL-based HVAC control strategies. By coupling the RL algorithm and an urban climate model with a BEM, the framework can investigate the impacts of RL-based HVAC control on both indoor climate and urban local climate. Simulations across cities with different climatic conditions were also used to explore the variations in these impacts, underscoring the importance of background climate when applying RL-based HVAC control technology. The framework of this study can be further adapted to evaluate the potential global-scale benefits of optimized HVAC strategies and their impact on the global climate. 

\section{Model and Methods}

\subsection{Community Land Model Urban} \label{sec:CLMU}

This study employs the Community Land Model Urban (CLMU), a process-based urban climate model that couples a BEM to simulate the building energy consumption and interactions between indoor climate and the local urban climate \citep{OLESONFEDDEMA20J.Adv.Model.EarthSyst.b}. As a component of the Community Earth System Model (CESM), CLMU-BEM has been thoroughly validated by \citet{LIETAL24NatureClimateChange}, showing its strong capability in (1) reproducing global patterns of urban heating and air conditioning energy consumption, (2) capturing the local scale heating and air conditioning energy consumption sensitivity to temperature variations, and (3) estimating urban air temperature responses to anthropogenic heat emissions. In addition, as coupled to an Earth system model, CLMU enables investigations on the long-term climate impacts of land-atmosphere coupling under the RL-based HVAC control. 

Based on the urban canyon concept \citep{OKE87a}, CLMU parameterizes urban areas with parameters for roofs, walls (both shaded and sunlit), and pervious or impervious roads. This model accounts for radiation trapping, heat conduction, and the hydrological processes on different surfaces, providing insights into how urban heat and moisture change. CLMU integrates a BEM that considers heat conduction, convection, radiation exchange, and ventilation \citep{OLESONFEDDEMA20J.Adv.Model.EarthSyst.b}. CLMU has become a useful tool in urban climate research \citep{OKE87a, ZHAOETAL21NatureClimateChange, ZHENGETAL21NatureCommunications}. More details of CLMU can be found in \citet{OLESONETAL10}. The urban surface data for CLMU simulations was extracted from \cite{JACKSONETAL10Ann.Assoc.Am.Geogr.c}. CLMU can compute the energy and turbulent fluxes based on the developed urban physical equations \citep{ZHENGETAL21NatureCommunications}. 
In particular, CLMU has been evaluated by both in-situ and remote sensing observations \citep{ZHAOETAL14Natureb, DEMUZEREETAL17Q.J.R.Meteorol.Soc.a, ZHAOETAL21NatureClimateChange, FITRIAETAL19ScientificReports, MOHAMMADHARMAYCHOI23Sustain.CitiesSoc.}, and compared with Weather Research and Forecasting (WRF) model simulation \citep{ZHAOETAL21NatureClimateChange}, which has been demonstrated to be suitable for urban climate simulations. 

\subsection{Building energy model} \label{sec:bem}

The BEM employed in CLMU primarily simulates (1) conduction of heat through building surfaces, including the roof, sunlit and shaded walls, and floor, (2) convection (sensible heat exchange) between interior surfaces and indoor air, (3) longwave radiation exchange between interior surfaces, and (4) ventilation encompassing natural infiltration and exfiltration \citep{OLESONFEDDEMA20J.Adv.Model.EarthSyst.b}. Energy conservation within the model is governed by the following equations~\ref{eq:ebalance1}--\ref{eq:ebalance5}: 

\begin{equation}\label{eq:ebalance1}
F_{rd,roof}+F_{cv,roof}+F_{cd,roof}=0,
\end{equation}
\begin{equation}\label{eq:ebalance2}
F_{rd,sumw}+F_{cv,sunw}+F_{cd,sunw}=0,
\end{equation}
\begin{equation}\label{eq:ebalance3}
F_{rd,shdw}+F_{cv,shdw}+F_{cd,shdw}=0,
\end{equation}
\begin{equation}\label{eq:ebalance4}
F_{rd,floor}+F_{cv,floor}+F_{cd,floor}=0,
\end{equation}
\begin{equation}\label{eq:ebalance5}
V_{B}\rho C_{p}\:\frac{\partial T_{iB}}{\partial t}-\sum_{sfc}A_{sfc}h_{c\nu,sfc}\big(T_{ig,sfc}-T_{iB}\big)-\dot{V}_{vent}\rho C_{p}(T_{ac}-T_{iB})=0,
\end{equation}

\noindent
where $F_{rd}$ represents the net longwave radiation flux (W m$^{-2}$), $F_{cv}$ is the convection flux (sensible heat flux), and $F_{cd}$ refers to the heat conduction flux (W m$^{-2}$) for each surface. $V_B$ denotes the volume of building air (m$^3$), $\rho$ is the density of dry air calculated at standard pressure $P_{std}$, and indoor air temperature $T_{iB}$ ($\rho = P_{std} / R_{da} T_{iB}$). $A_{sfc}$ represents the surface area (m$^2$), $h_{cv, sfc}$ is the convective heat transfer coefficient (W m$^{-2}$ K$^{-1}$), and $T_{ig, sfc}$ is the interior surface temperature of each surface, where the subscript $sfc$ represents either the roof, sunlit wall, shaded wall, or floor. In addition, $\dot{V}_{vent}$ is the ventilation airflow rate (m$^3$ s$^{-1}$, an adjustable parameter) and $T_{ac}$ is the urban canopy air temperature (K). 

By solving these equations, the interior surface temperature and indoor air temperature can be determined. Subsequently, by analyzing the thresholds of maximum and minimum indoor air temperatures, further constraints on indoor air temperature can be established. This allows for the calculation of the energy required for heating and cooling to maintain desired indoor temperature conditions. The heating and cooling fluxes are derived from the following:  
  
\begin{equation}\label{eq:cool}
F_{cool}=\begin{cases} 
    \frac{H\rho C_{p}}{\Delta t}\left(T_{iB}^{t+1}-T_{max}\right) , & \text{if } T_{iB}^{t+1}>T_{max} \\ 
    0 , & \text{if } T_{iB}^{t+1}<=T_{max} 
    \end{cases}, 
\end{equation}

\begin{equation}\label{eq:heat}
F_{heat}=\begin{cases}
\frac{H\rho C_{p}}{\Delta t}\left(T_{min}-T_{iB}^{t+1}\right) , & \text{if }T_{iB}^{t+1}<T_{min}\\
    0 , & \text{if } T_{iB}^{t+1}>=T_{max} 
    \end{cases},
\end{equation}

\noindent
where $F_{cool}$ and $F_{heat}$ are the fluxes for cooling and heating (W m$^{-2}$), respectively. The waste heat produced by air conditioning and heating can be released to the urban canyon floor. The amount of waste heat is defined as 0.6$F_{cool}$ and 0.2$F_{heat}$ \citep{OLESONFEDDEMA20J.Adv.Model.EarthSyst.b}. 

\subsection{Model configuration of CLMU} \label{sec:mc_clmu}

\begin{figure*} [htbp]
    \centering
    \includegraphics[width=\textwidth]{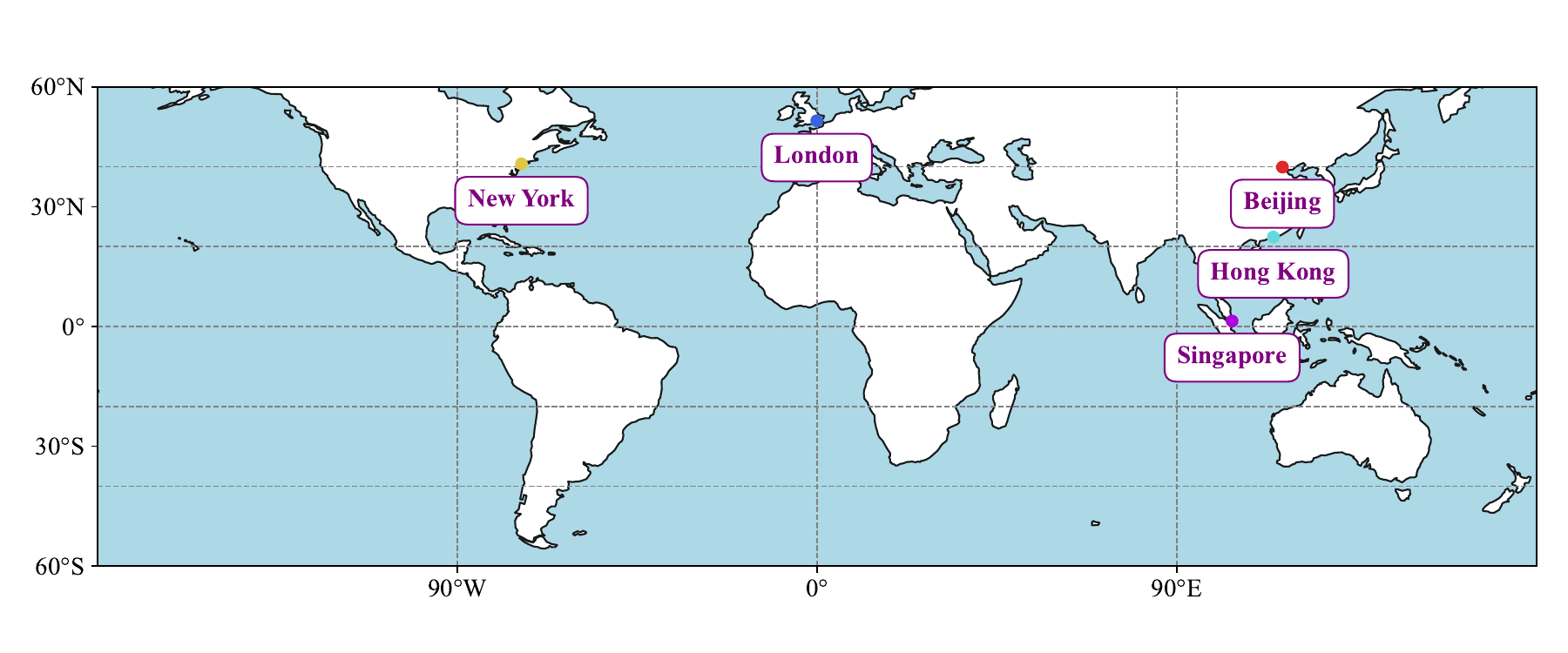}
    \caption{Cities selected for this study.}
    \label{fig:cities}
\end{figure*}

In this study, we used the observational forcing data (2002--2013) of UK-Kin site from the Urban-PLUMBER Phase 2 project (\url{https://urban-plumber.github.io/UK-KingsCollege/}) in London, UK. The forcing data of New York, Beijing, Hong Kong, and Singapore were derived from the single-level fifth-generation European Centre for Medium-Range Weather Forecasts reanalysis (ERA5) covering the period from 2011 to 2023. The urban surface parameters were adopted from the urban surface dataset (\url{https://svn-ccsm-inputdata.cgd.ucar.edu/trunk/inputdata/lnd/clm2/rawdata/mksrf_urban_0.05x0.05_simyr2000.c170724.nc}) and soil texture dataset (\url{https://svn-ccsm-inputdata.cgd.ucar.edu/trunk/inputdata/lnd/clm2/rawdata/mksrf_soitex.10level.c010119.nc}) from CESM. All simulations are run at a 12-year length, and the first 10 years are for the spin-up stage (to generate the initial condition for running the simulations) and a further 2-year simulation is for the analysis. The selected cities are the most highly developed cities in the world (based on 2024 Globalization and World Cities Research Network, GaWC) and have different background climates. The locations of the selected cities are shown in Figure~\ref{fig:cities}, which covers different latitudes. The forcing temperatures of each city are displayed in Figure~S1, showing the difference in their background temperature coverage. All the simulations were performed using Pyclmuapp---a Python package to streamline the CLMU simulations \citep{YUETAL25Environ.Model.Softw.}. 
The simulations using the default HVAC control strategies in CLMU served as baselines for comparing the RL strategies. The default AC set points for London, New York, Beijing, Hong Kong, and Singapore are 380.00 K, 310.00 K, 310.00 K, 310.10 K, and 380.00 K, respectively, indicating that AC will not be active in London and Singapore; the default heating set points are 290.10 K, 285.10 K, 285.10 K, 290.10 K, and 285.10 K, respectively. The ventilation rate is set as a constant value of 0.3. 

\subsection{Reinforcement Learning} \label{sec:rl}

\begin{figure*} [htbp]
    \centering
    \includegraphics[width=\textwidth]{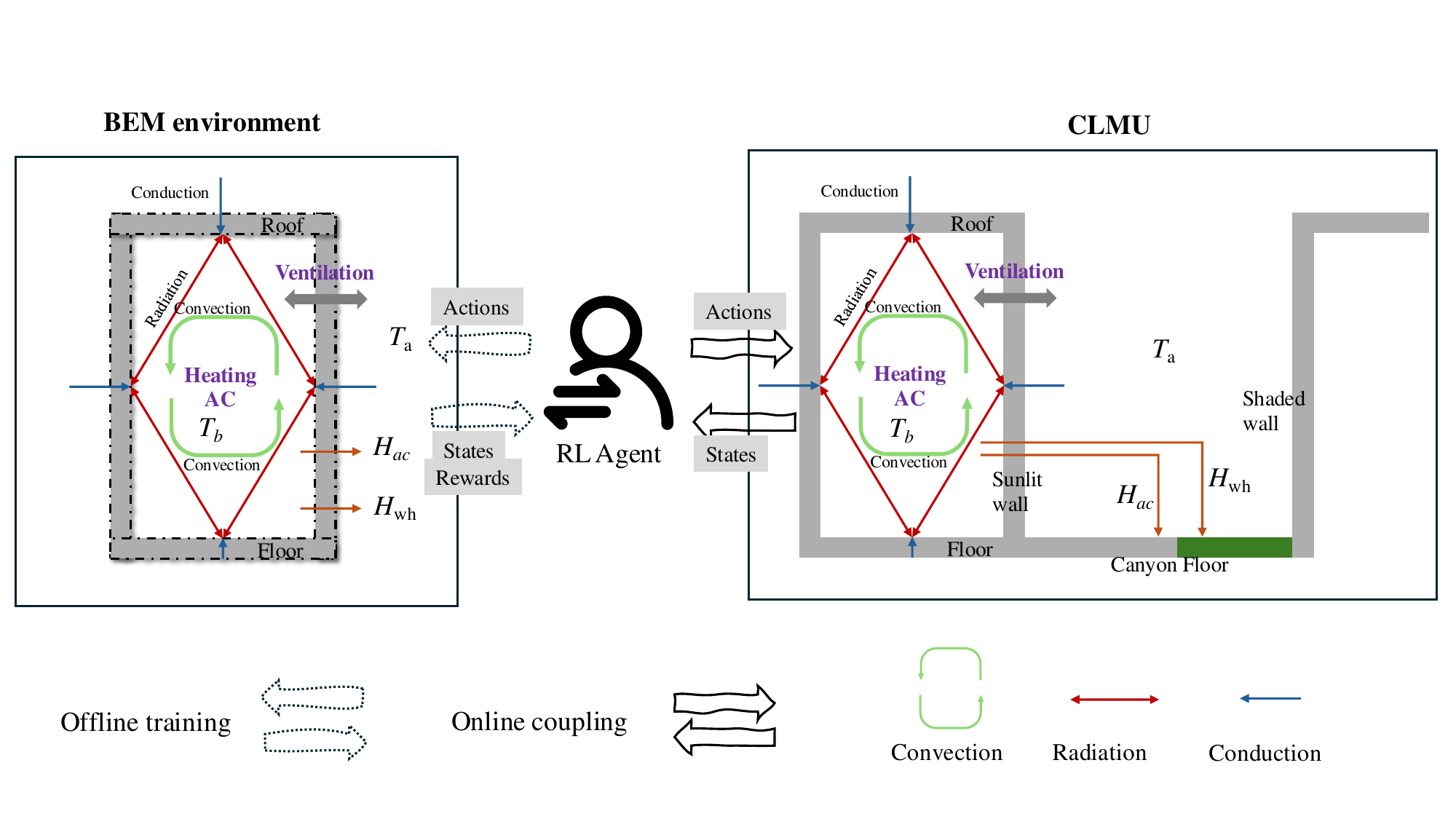}
    \caption{Schematic diagrams of RL training with surrogate and coupling with CLMU. $T_\text{a}$ is the urban canopy air temperature; $T_\text{b}$ is the building indoor air temperature; $H_\text{wh}$ is the heat flux due to inefficiencies in heating and air conditioning systems, as well as energy losses during the conversion of primary energy sources into usable end-use energy; $H_\text{ac}$ is heat flux extracted from the building's interior by the air conditioning system. Both $H_\text{wh}$ and $H_\text{ac}$ can be put into the canyon floor as sensible heat in CLMU simulations.}
    \label{fig:workflow}
\end{figure*}

The overall workflow of reinforcement learning includes problem definition, agent selection, optimization, and online coupling. Problem definition and description of the environment, action, states, and reward are detailed in Section~\ref{sssec:pd}--\ref{sssec:sr}, agent selection is described in Section~\ref{sssec:agent}, agent optimization is presented in Section~\ref{sssec:opt}, and agent online coupling with CLMU is in Section~\ref{sssec:online}. The framework schematic diagram of RL training with surrogate and coupling with CLMU is shown in Figure~\ref{fig:workflow}.

\subsubsection{The fundamentals of RL}\label{sssec:bd}
The foundational concepts of RL are introduced in this section. 

\textbf{Environment} is the interactive system that can interact with the actions from the agent. It provides the agent with information about its current state, responds to the agent's actions, and generates rewards or penalties based on actions. 

\textbf{Agent} is the decision-maker that interacts with the environment. It produces decisions based on the state. The agent makes the decision using the policy function ($\pi$). $\pi$ can be a mapping ($\pi:S \rightarrow A$) or a probability distribution ($\pi(a|s) = P(A_t=a|S_t=S)$), which are applied for deterministic or stochastic action, respectively. 

\textbf{State} is a snapshot from the environment that influences the decision of the agent. The states are the subset of the state space (all possible states), which can be continuous or discrete. 

\textbf{Action} is the decision made by the RL agent. The action is determined by the current state, regardless of state history. Action can be continuous or discrete, depending on the problem definition. 

\textbf{Episode} refers to a complete sequence of interactions between the agent and the environment, starting from an initial state and continuing until a terminal state is reached. The goal of the agent is to maximize its cumulative reward during the episode. 

\textbf{Reward} represents how well the action is in a given state. It is a function of action and state. The sequential problem focuses on the long-term reward (return), which is called return (cumulative reward) with a discount factor ($\gamma$). The return is given by: 
\begin{equation}
    r_{t} = R_{t} + \gamma R_{t+1} + \gamma^{2} R_{t+2} + ...
\label{eq:return}
\end{equation}
\noindent
where $R_{t}$ is the reward obtained at time $t$. The return $r_t$ is the cumulative reward with a discount factor $\gamma$ from time $t$ to the end of the episode. 

\subsubsection{Problem formulation}\label{sssec:pd}

CLMU accounts for the trapped radiation, surface energy balance, and the BEM. By specifying atmospheric forcing, the urban climate model enables a functional relationship between specific urban parameters and corresponding climate responses. With active control of building energy strategies, such as HVAC systems, both energy consumption and thermal comfort can be affected. This forms the basis for an optimization problem aimed at reducing energy consumption and enhancing thermal comfort through HVAC control. In this context, the objective functions incorporate both energy consumption and thermal comfort, with HVAC control strategies serving as the decision variables. 

HVAC control is a dynamic, multi-step problem, making it challenging to solve with traditional optimization algorithms. Therefore, we employ RL to tackle this problem. RL falls into a special class of optimization, aiming to achieve optimal sequential decision-making by constructing one or more agents. Like traditional optimization problems, RL requires an explicit definition of the problem. For HVAC control problems, the agent will consider the current local urban climate and indoor climate and make decisions on HVAC control strategies, and states will be updated according to the decision. 

\subsubsection{Environment for RL}\label{sssec:env}

CLMU is developed in Fortran, a language well-suited to numerical modeling. However, the compatibility of Fortran with ML and RL is limited, as most RL frameworks---especially DRL frameworks---are implemented in Python. This discrepancy poses challenges for coupling RL with the Fortran-based model. Consequently, a surrogate model, e.g., an ML model or a simplified high-fidelity model, is often applied to replace the complex model, facilitating integration with RL frameworks. 

In this study, the Fortran-based BEM within the CLMU was ported to Python, enabling the construction of an RL-compatible learning environment. In particular, this Python version functions as an offline BEM, meaning that the HVAC system cannot affect the local urban climate. This limitation will be further discussed in Section~\ref{sssec:sr}. The inputs, outputs, and other details of Python version BEM are described in Table~S1--Table~S2. The validation of Python version BEM is provided in Figure~S2--Figure~S6, showing an error on the order of 10$^{-5}$ K in building air temperature calculation, which should be the rounding error in floating-point numbers due to differences in algorithms, precision settings, or hardware architectures.

It is important to note that the Python version of BEM is used solely as a surrogate model to construct the RL environment and train the agent. The performance of the RL model is evaluated based on results obtained from the online coupling with CLMU (Section~\ref{sssec:online}). We established a Gym environment interface---a popular RL framework (Copyright \textcopyright~2023 Farama Foundation)---to integrate the Python-based BEM reinforcement learning environment. Each environment episode spanned 17,520 steps, equivalent to one year of the BEM (or CLMU) simulations. The training period was set to 2013 for London, while for other cities (New York, Beijing, Hong Kong, and Singapore), it was 2023. Model results were evaluated for 2012 in London and 2022 in the other cities. 

\subsubsection{Action}\label{sssec:action}

To implement HVAC control strategies, the agent dynamically adjusts the indoor maximum temperature, minimum temperature, and ventilation efficiency, corresponding to the AC set point temperature, heating set point temperature, and ventilation intensity, respectively. The action spaces are defined as follows: AC set point temperature (25--35°C), heating set point temperature (10--20°C), and ventilation intensity (0.3--0.5). The action can be continuous or discrete according to algorithms, and the final action space is determined by the algorithms described in Section~\ref{sssec:agent}. 

\subsubsection{States and reward}\label{sssec:sr}

A state is defined by values of the AC set point temperature, heating set point temperature, ventilation intensity, indoor air temperature, and local urban air temperature (the canopy air temperature). While states are affected by actions, it should be noted the local urban air temperature remains unaffected by actions in the surrogate BEM. 

The reward function is multifaceted, integrating both an energy term and a thermal discomfort adapted from `Sinergym' \citep{JIMENEZ-RABOSOETAL21Proc.8thACMInt.Conf.Syst.Energy-Effic.Build.CitiesTransp.}, as follows:
\begin{equation}
    R = -w \lambda_P (P_{\text{ac}} + P_{\text{heat}}) - (1-w) \lambda_T \left(|t_{\text{building}} - T_{\min}| + |T_{\max}-t_{\text{building}}|\right)
\label{eq:reward}
\end{equation}
where $P_{\text{ac}}$ and $P_{\text{heat}}$ represent the adjusted energy demand for air conditioning and heating. These values are calculated from the air conditioning and heating heat fluxes obtained from CLMU/BEM, divided by their respective coefficient of performance (COP) and power plant efficiency (Peff). In this study, the COP and Peff for air conditioning are set to 0.9 and 0.96, respectively, while for heating, they are set to 3.6 and 0.43. These values are identical to CLMU default setting. $\lambda_P$ and $\lambda_T$ are the constants for scaling the magnitudes of energy and comfort terms, respectively, both of which are set to 1 in this study. Here, $t_{\text{building}}$, $T_{\min}$, and $T_{\max}$ denote the indoor air temperature, the minimum comfortable temperature, and the maximum comfortable temperature, respectively. In this study, $T_{\min}$ and $T_{\max}$ are set to 18°C and 24°C. The weight ($w$) for the energy consumption and thermal discomfort terms is set to 0.1. The return ($r$) is given by equation~\ref{eq:return}. 

\subsubsection{Agent}\label{sssec:agent}

Q-learning \citep{WATKINSDAYAN92MachineLearning}, deep Q learning (DQN) \citep{MNIHETAL15Naturea}, and soft actor-critic (SAC) \citep{HAARNOJAETAL18a} were selected to train the RL agents. The selected methods are off-policy algorithms, meaning the target policy is different from the policy used to collect data from the environment for training the policy. Thus, they can leverage collected data to train the policy, and can benefit from the ``replay buffer''---an experience storage mechanism used to retain collected data for learning. However, these methods differ in complexity. Here, these three methods were selected to explore how the algorithm complexity influences HVAC control performance. The details of these methods are provided in the Supplementary Information; here, we briefly introduce the algorithms. 

Q-learning is a typical value-based method, which learns the Q-function (action-state value function, $Q(s,a)$). The details of Q-learning are described in Section~S1.2. The agent selects actions based on the learned Q-function, aiming to maximize the values of action. The Q-function is presented as a table storing the action, state, and action-state value. Therefore, action and state must be discretized.

DQN is a representative DRL algorithm, which is derived from Q-learning. DQN employs a neural model to parameterize the Q-function ($Q_\theta(s,a)$ where $\theta$ is the parameters of neural networks). The DQN can thus map the continuous and discretized state to the action-state value. The details of DQN are presented in Section~S1.2. In neural network training, the loss with parameter $\theta$ is the objective. $\theta$ can be optimized using the descent gradient. DQN algorithms train two networks simultaneously, the target network and the master network. The target network is updated only after the master network has been updated several times, which ensures stable convergence of the networks \citep{MNIHETAL15Naturea}. Usually, the samples generated by the training process are put into a replay buffer for sample reuse and correlation breaking among experiences. These tricks enable stable and efficient training for DQN. However, DQN is usually suitable for discrete actions (finite action) because its policies are given by $argmax_a(Q(s,a))$. This calculation will become exhaustive when the action is infinite.

SAC is a DRL algorithm designed to achieve sample-efficient learning while leveraging the benefits of entropy maximization for enhanced exploration and stability \citep{HAARNOJAETAL18a}. The details of SAC can be found in Section~S1.3. Entropy refers to a measure of randomness or uncertainty in the policy's action selection, which can be used to encourage exploration and improve the stability of learning. The entropy constraint allows the SAC to explore a wider range of possible actions rather than selecting deterministic actions during each training. In practice, the degree of exploration of SAC is controlled by a temperature factor (coefficient of entropy), which is equal to the deterministic strategy when it is set to zero. This approach makes SAC training fast and stable \citep{HAARNOJAETAL18a}. SAC can handle both continuous states and actions.

Due to the different characteristics of these three algorithms, specific processing to the actions and states was applied accordingly. For Q-learning, both actions and states are discretized, while for DQN, only actions are discretized. The SAC uses continuous action and state spaces. To simplify, the discrete action space is reduced to 8 categories. We defined the AC set point at two discrete values, 26°C (299.15 K) and 55°C (328.15 K), the heat set point at 15°C (288.15 K) and -15°C (258.15 K), and the ventilation intensity at 0.3 and 0.5. By combining these values, 8 action types were created, which keep the action space compact, facilitating the RL training process. As Q-learning is performed with the discretized state, we used linear combination encoding for discretization. Specifically, we scaled the ventilation rate by 10, rounded all state variables to integers, and then mapped the discrete states to a unique index using a weighted sum, where each state variable was multiplied by a specific power of 10 to ensure uniqueness in encoding.

\subsubsection{Optimization}\label{sssec:opt}

In the RL optimization processes, the reward discount factor was set to 0.99. A total of 50 episodes were performed during training in a total time step of $8.76 \times 10^5$. The DQN and SAC algorithms were adapted from the CleanRL repository (\url{https://github.com/vwxyzjn/cleanrl}) \citep{HUANGETAL22J.Mach.Learn.Res.} to meet the specific requirements of this study. Unless otherwise noted, all algorithm parameters followed the default CleanRL settings to ensure consistency across experiments conducted for different cities. The algorithm details, network architectures, training steps and other parameters are described in Section~S1. The training curves for different RL algorithms are shown in Figure~S7. 

\subsubsection{Coupling DRL with CLMU} \label{sssec:online}

During training, DRL consists of multiple neural networks, but only the policy network is used when coupling RL with CLMU. The policy network is a two-layer perception with a simple architecture (details in Section~S1.3), relying primarily on linear matrix operations that are compatible with Fortran computations. The inference was built by saving the neural network weights and biases and leveraging the built-in `matmul' function of Fortran, which performs matrix multiplication on numeric arguments, to execute the neural network calculations. The inference process, including network parameters loading and action sampling, is implemented in a module named `\url{sac_actor.f90}'. This module is called within the BEM of CLMU (in the `\url{UrbBuildTempOleson2015Mod.F90}' file of CLMU). In each step, subroutines in `\url{sac_actor.90}' are called, and the AC set point temperature, heating set point temperature, and ventilation intensity are updated based on current states. BEM then proceeds with its calculations as usual and outputs results for the next time step. 

\section{Results and discussion}

\subsection{Performance of RL algorithms}

This study tested the performance of three different RL algorithms for HVAC control across five cities with various background climates. SAC algorithm achieves a higher episode return with fewer training steps and converges more quickly within the steps (Figure~S7). In contrast, both DQN and Q-learning exhibit lower learning efficiency and slower convergence compared to SAC (Figure~S7). The SAC algorithm also achieved the highest mean reward among all cases, indicating the best testing results (Table~\ref{tab:rl_perf}). Since SAC is an algorithm that can be used for continuous control \citep{HAARNOJAETAL18a}, it may benefit HVAC control more than DQN and Q-learning. Consequently, SAC was identified as the optimal algorithm under the given experimental conditions, and subsequent analyses are based on SAC results.

\begin{table*}[htbp]
\caption{The mean reward of different algorithms across cities. Results are based on three epochs of testing in the surrogate environment. A higher value indicates better performance.}
\centering
\begin{tabular}{l|cc|cc|cc}
\hline
City      & \multicolumn{2}{c|}{DQN} & \multicolumn{2}{c|}{Q-Learning} & \multicolumn{2}{c}{SAC} \\
                   & Mean       & Std            & Mean       & Std       & Mean       & Std       \\
\hline
London             & -11.12              & 0.84              & -12.10              & 0.01              & \textbf{-7.65}               & 0.00               \\
New York           & -13.20              & 0.18              & -12.50              & 0.01              & \textbf{-8.71}               & 0.00               \\
Beijing            & -11.32              & 0.55              & -13.02              & 0.01              & \textbf{-7.97}               & 0.00               \\
Hong Kong          & -8.18               & 0.01              & -8.73               & 0.02              & \textbf{-6.92}               & 0.00               \\
Singapore          & -9.30               & 0.00              & -10.47              & 0.03              & \textbf{-7.66}               & 0.00               \\
\hline
\end{tabular}
\label{tab:rl_perf}
\end{table*}

\subsection{Policy evaluation by coupling RL and CLMU}

\begin{figure*} [htbp]
    \centering
    \includegraphics[width=\textwidth]{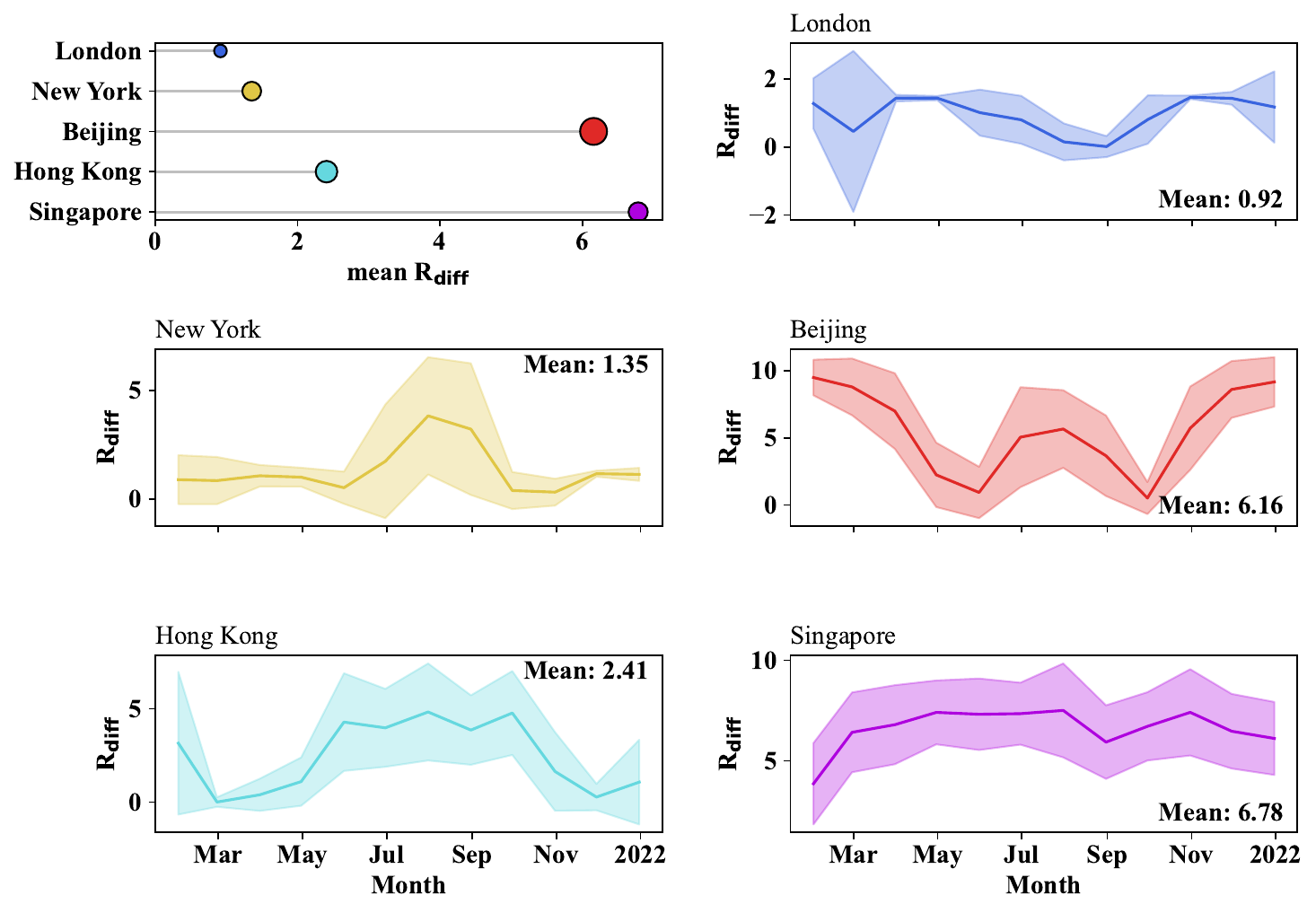}
    \caption{Mean reward difference in RL case and baselines (CLMU default setting) across cities and monthly profiles. The size of the marker indicates the standard error.}
    \label{fig:online_val}
\end{figure*}

Although training RL models using the high fidelity surrogate eliminates the need to run a full computational model \citep{GONGDUAN17EnvironmentalModelling&Softwareb, GONGETAL16WaterResour.Res.a, SUNETAL22EnvironmentalModelling&Softwarec}, it inevitably introduces some loss of accuracy and fails to capture the two-way interaction between indoor climate and local urban climate. Therefore, instead of using the surrogate model, we embed the RL model into CLMU for evaluations. The RL model is used to adjust the AC set point, heating set point, and ventilation based on input states, while the BEM in CLMU subsequently computes the system response using these updated parameters. To evaluate the performance of RL models, rewards from RL-based control are compared to baselines from the default setting of CLMU, where no intelligent HVAC control is applied (Section~\ref{sec:mc_clmu}). The results show that RL achieved higher rewards than the baselines, demonstrating the advantage of RL in HVAC control (Figure~\ref{fig:online_val}).

The reward differences (R$_\text{diff}$) in RL and default cases varied across geographic locations. R$_\text{diff}$ for London, New York, Beijing, Hong Kong, and Singapore are 0.92, 1.35, 6.16, 2.41, and 6.76, respectively (Figure~\ref{fig:online_val}). The R$_\text{diff}$ of London and New York---cities with colder climates---are lower than those of cities with warmer climates, like Hong Kong and Singapore. In other words, the benefits of applying RL for HVAC control are markedly influenced by geographic location, with relatively lower rewards in colder cities compared to those with warmer climates. 

R$_\text{diff}$ fluctuates seasonally (Figure~\ref{fig:online_val}), which can be attributed to the dynamic background climate in different cities. The monthly average R$_\text{diff}$ of London remained relatively steady, varying between 0 and 2 with less fluctuation. Cities like New York, Beijing, and Hong Kong displayed more pronounced monthly variations, with notably higher rewards during the summer months. These variations demonstrate the importance of considering regional climate diversity when applying RL to HVAC control because both seasonal climate and background climate can influence the effectiveness of RL-based strategies. This result reinforces that the uncertainty of climatic factors can directly affect the efficiency of HVAC systems \citep{BAZAZZADEHETAL25EnergyReports}. 

RL is effective during cold and hot seasons---when HVAC systems are frequently in use  (Figure~\ref{fig:online_val}). This finding strongly supports the effectiveness of RL-based HVAC control in optimizing HVAC operations. The background climate determines when the HVAC system is activated in most cases, which in turn affects the effectiveness of the RL strategy. For instance, heating systems are typically used in cold climates or during winter, while air conditioning is more prevalent in hot climates or summer, and thus RL gets more reward in these conditions (Figure~\ref{fig:online_val}). During comfortable seasons like spring and autumn, HVAC systems are often turned off and the reward of RL is lower. With global warming, some cities will significantly increase the use of AC, which may further increase the value of HVAC control systems. To better assess the adaptability of RL strategies on different background climates, we recommend investigating the background climate and HVAC usage patterns, as this will provide valuable insights into estimating the overall benefits of the strategy. 

\subsection{The impact of RL strategies on indoor climate and local urban climate}

The indoor climate interacts with the local urban climate through heat emission and ventilation, while the local urban climate can also influence the indoor thermal environment. Changes in the indoor thermal state can thus have a cascading impact on the local urban climate. This study quantified the complex effects of RL-based control strategies using results from the online coupling approach. 

\begin{figure*} [htbp]
    \centering
    \includegraphics[width=\textwidth]{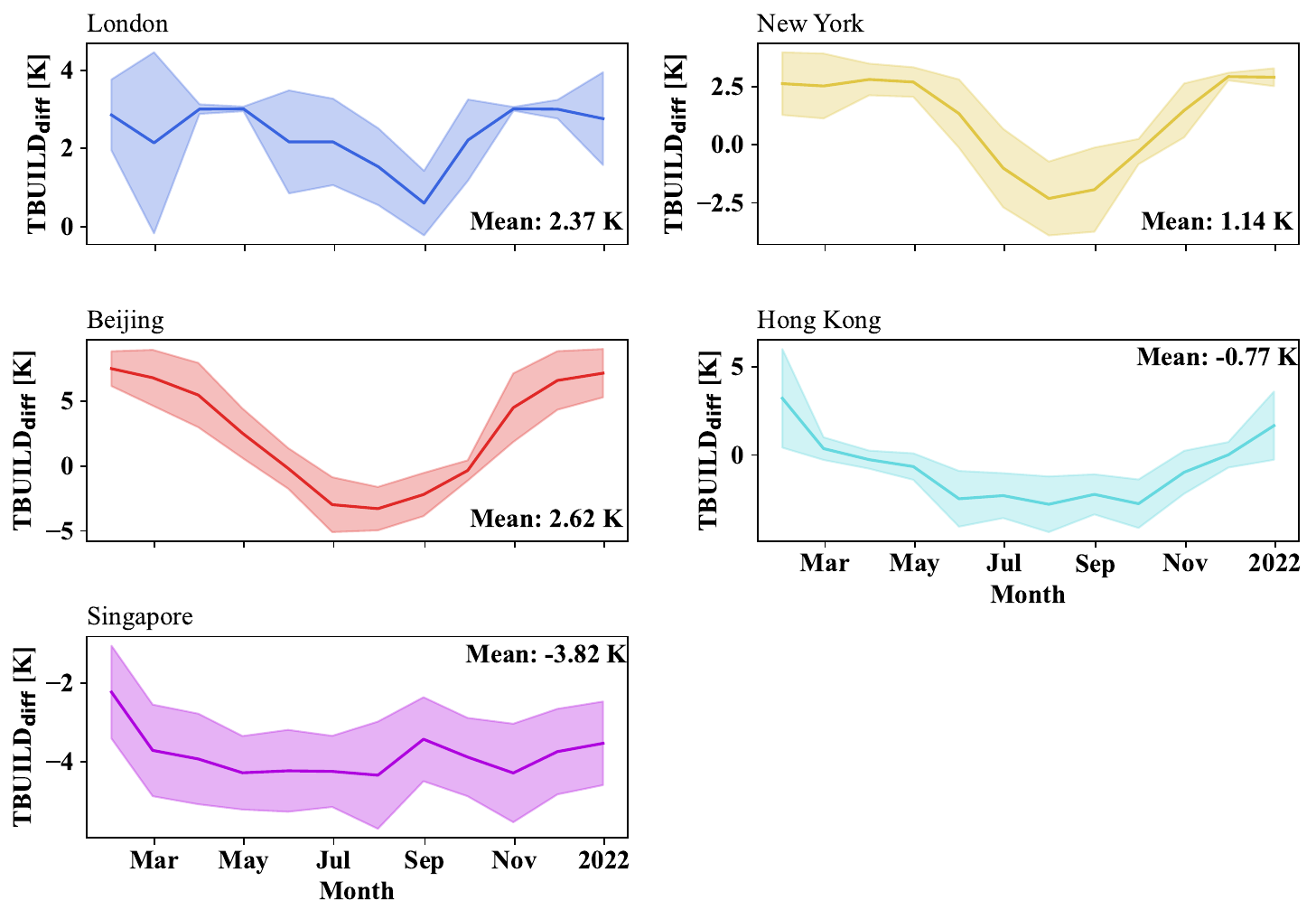}
    \caption{Monthly profile of difference of indoor air temperatures of buildings (TBUILD$_\text{diff}$) in RL case and CLMU default case across cities.}
    \label{fig:tbuild}
\end{figure*}

The impact of these strategies on indoor air temperatures of buildings (TBUILD) was investigated and shown in Figure~\ref{fig:tbuild}. The changes of TBUILD in RL and default cases (TBUILD$_\text{diff}$) across London, New York, Beijing, Hong Kong, and Singapore varied, with average TBUILD$_\text{diff}$ of 2.37, 1.14, 2.62, -0.77, and -3.82 K, respectively. This highlights a clear influence of background climate on TBUILD$_\text{diff}$. For example, tropical cities like Hong Kong and Singapore had negative TBUILD$_\text{diff}$, while other cities exhibited positive values, indicating that the impacts of RL control strategies on indoor climate vary distinctly across different background climates. New York, Beijing, and Hong Kong demonstrate notable seasonal variations, where TBUILD generally increased in winter and decreased during warmer months. London and Singapore show less variation across different seasons. These results indicate that RL control strategies effectively increased TBUILD under cold climates while lowering TBUILD under warmer climates. One possible explanation is that the reward function prioritized thermal comfort, steering RL strategies toward creating a more comfortable indoor environment.

\begin{figure*} [htbp]
    \centering
    \includegraphics[width=\textwidth]{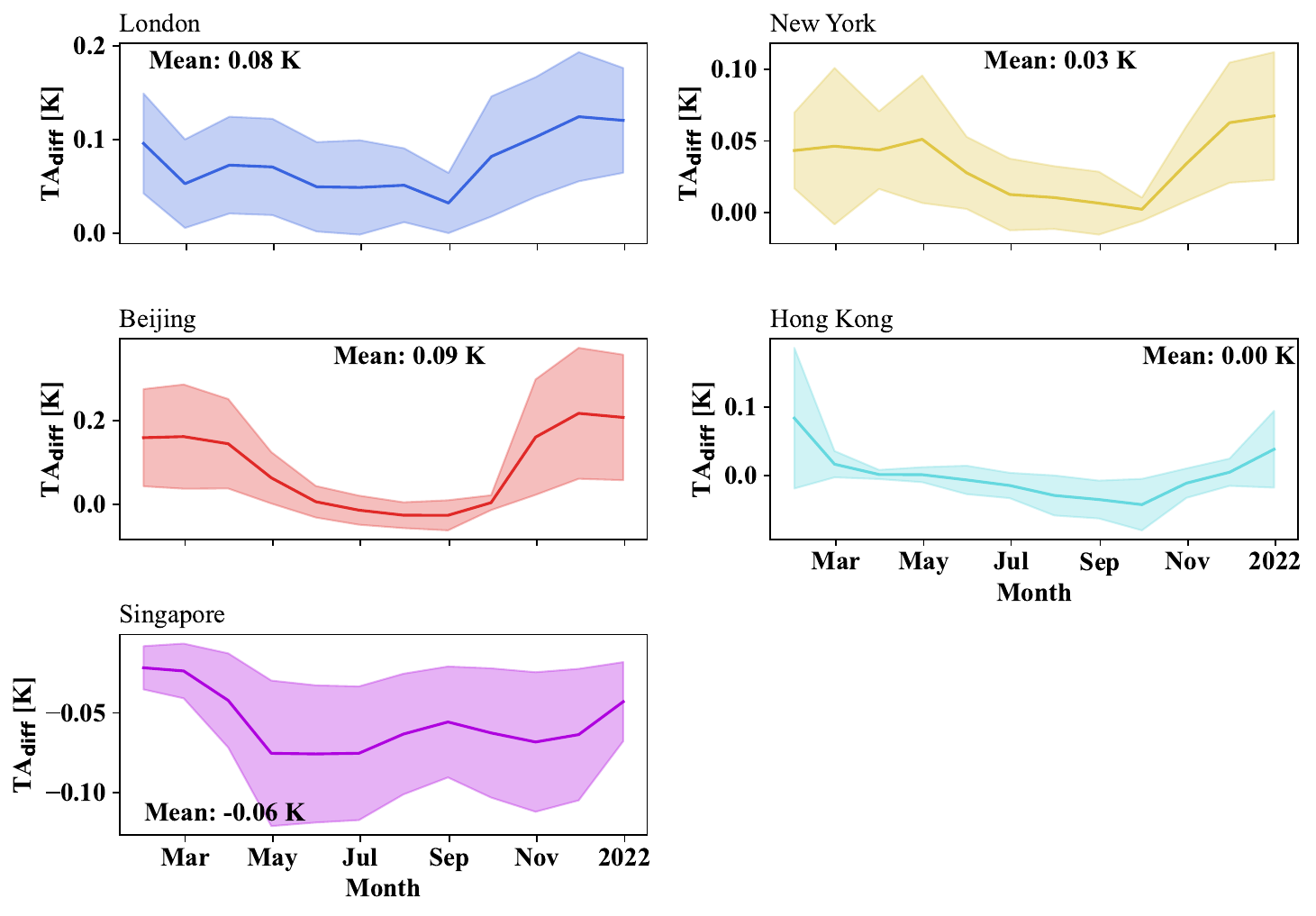}
    \caption{Monthly profile of outdoor air temperature difference (TA$_\text{diff}$) in RL case and CLMU default case across cities.}
    \label{fig:tsa}
\end{figure*}

In London, New York, and Beijing, outdoor air temperatures (TA; equal to urban canyon air temperature) increased after implementing RL strategies, with mean TA differences in RL and default cases (TA$_\text{diff}$) of 0.08, 0.03, and 0.09 K, respectively (Figure~\ref{fig:tsa}). For Hong Kong and Singapore, the RL strategies decrease the outdoor air temperature at most months, with mean TA$_\text{diff}$ of 0.00 and -0.06 K. Thus, RL strategies in colder regions tend to increase outdoor air temperatures, whereas in warmer regions, this trend may be reversed. The increase of TA was more pronounced in colder seasons (or less decrease in Singapore), while the increase of TA was smaller or decreased further during hotter periods (Figure~\ref{fig:tsa}). Changes in local urban humidity are closely related to temperature variations, typically exhibiting an opposite trend of TA (Figure~S8). Specifically, the average local urban humidity tends to increase in London, New York, and Beijing, but shows a decreasing trend in Hong Kong and Singapore (Figure~S8). It is important to note that these results were obtained from CLMU simulations in which land climate dynamics were simulated independently, without coupling to a global atmospheric model. When running a fully coupled CESM simulation, the small changes in land may lead to significant climate differences and potentially cause future shifts \citep{KAYETAL15Bull.Am.Meteorol.Soc.}. Future study should employ more comprehensive model simulations to analyze the large-scale impacts of RL strategies, particularly their effects on global climate. 

\begin{figure*} [htbp]
    \centering
    \includegraphics[width=\textwidth]{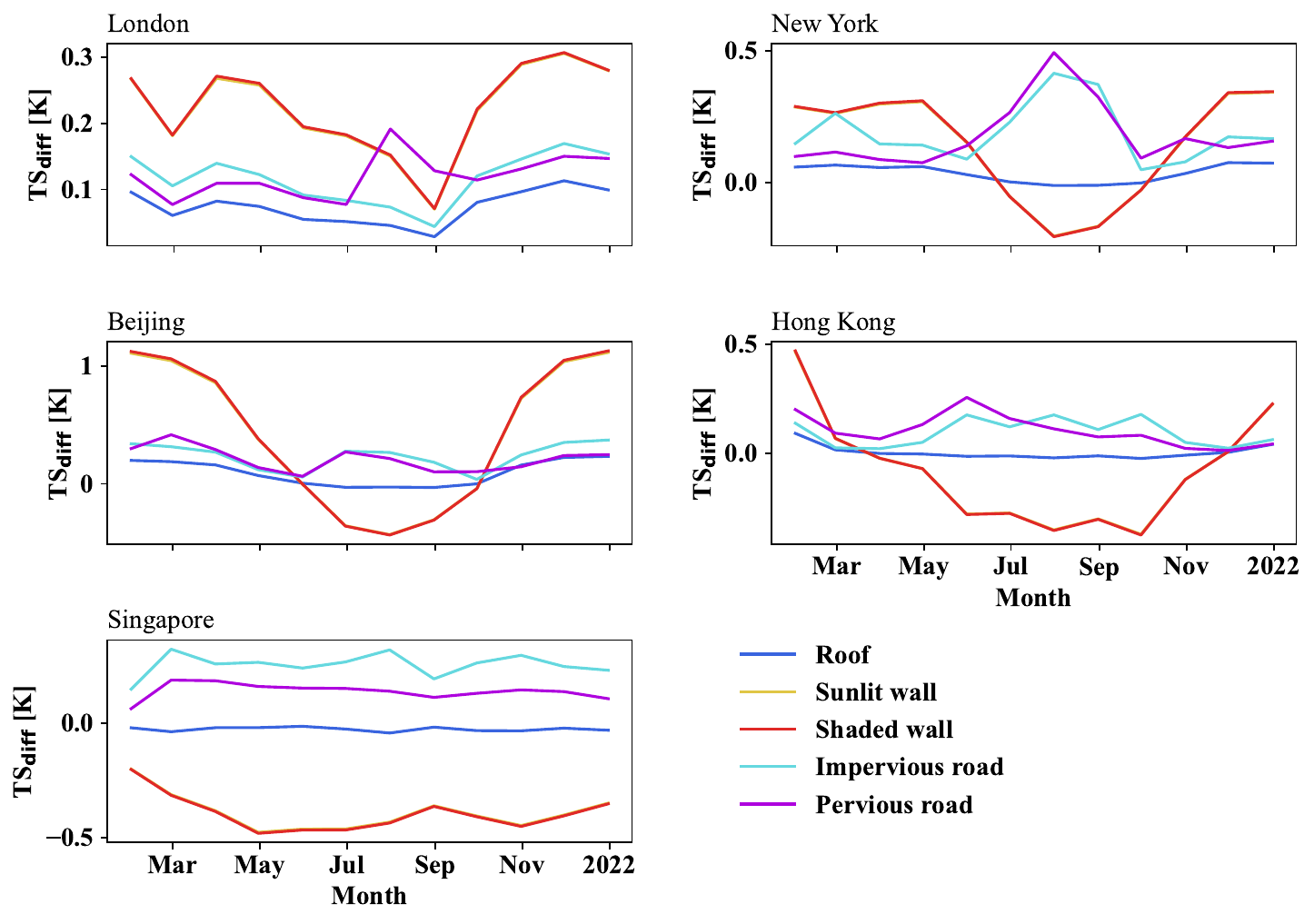}
    \caption{Monthly profile of surface temperature difference in RL case and CLMU default case across cities. Sunlit wall and shaded wall are identical. }
    \label{fig:tskin}
\end{figure*}

Results on temperature variations across different urban surfaces indicate that wall temperatures were the most sensitive to RL strategies implementation, whereas roof surface temperatures were the least affected (Figure~\ref{fig:tskin}). Additionally, a consistent increase in road surface temperature is shown. Waste heat, generated during the cooling and heating, was directly added to the canyon floor as sensible heat in CLMU \citep{OLESONFEDDEMA20J.Adv.Model.EarthSyst.b}, resulting in a persistent heat increase on the road surface. It is important to acknowledge that this direct addition method may introduce bias, potentially affecting the accuracy of local urban air temperature results.

Since the walls usually receive less radiation than the roof, their temperature variations are more closely related to indoor temperature changes. From the monthly temperature variation curves, only the surface temperature of walls shows a decrease after implementing RL strategies (Figure~\ref{fig:tskin}). Thus, the reduction in outdoor temperature during specific cities or seasons after implementing the RL strategy may be attributed to changes in wall temperatures (Figure~\ref{fig:tsa}). When the cooling effect of walls appears to outweigh the warming effect of other surfaces, the outdoor air temperatures may decrease. It is important to note that walls act as intermediaries between indoor and outdoor climates. As a result, temperature changes in walls induced by the indoor climate can influence outdoor climate, which in turn affects heat transfer processes and HVAC systems, making the overall dynamics potentially more complex than our current explanation. Additionally, CLMU is based on the urban canyon concept, which idealizes urban environments and buildings. We acknowledge the potential errors in climate and energy estimations that arise from this simplified modeling approach.

\subsection{Reward responses to weight modifications in a fixed RL policy framework} 

\begin{figure*} [htbp]
    \centering
    \includegraphics[width=\textwidth]{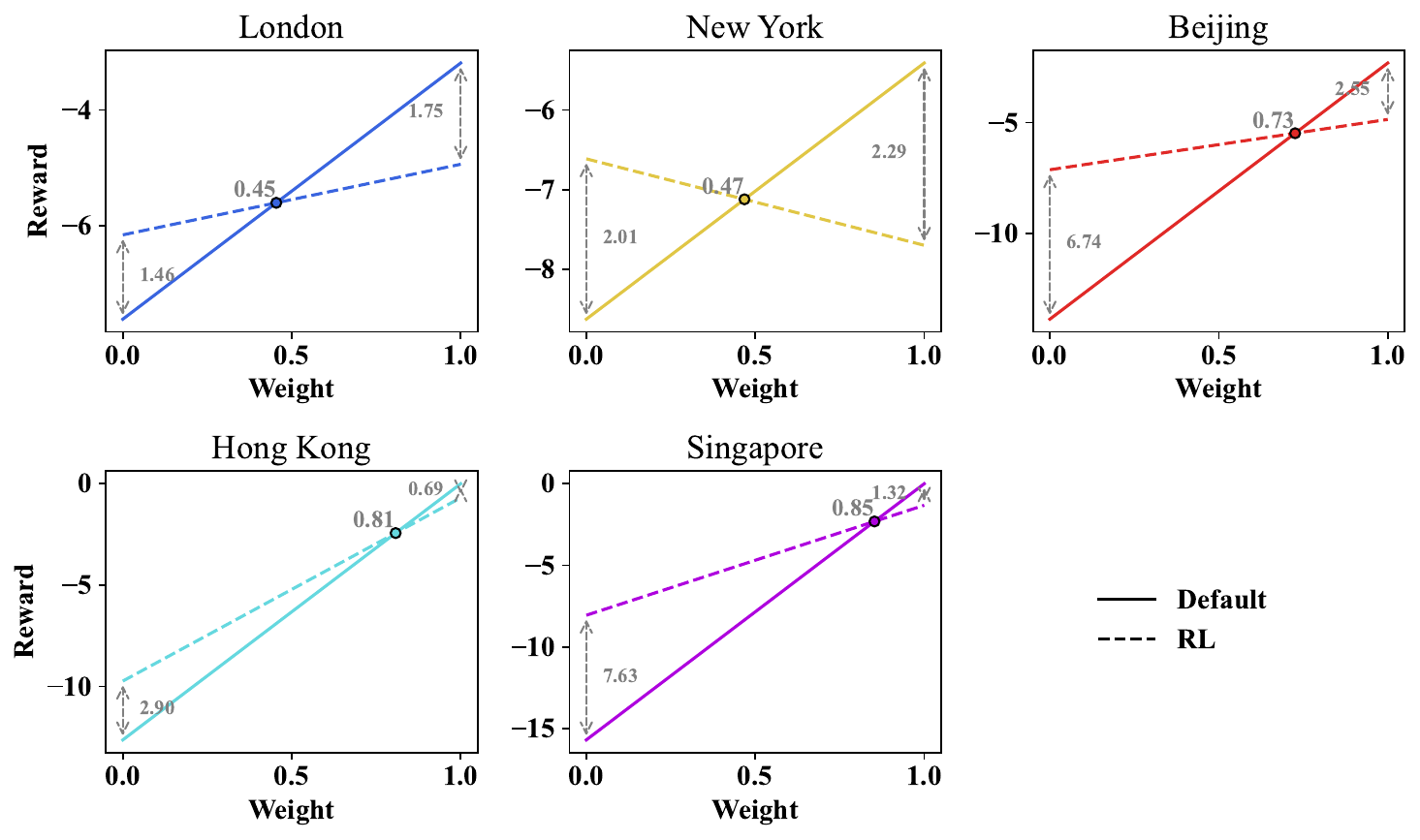}
    \caption{Impacts of reward weight modifications on reward response in a fixed RL policy (trained with $w$ = 0.1), compared to baselines across cities. }
    \label{fig:w_sens}
\end{figure*}

The RL reward function was a linearly weighted function in this study, encompassing both the energy term and thermal comfort. These two objectives were connected by a weight ($w$), set to 0.1 during training. A higher weight results in a larger contribution of the energy term to the total reward. This weight may vary in real-world applications depending on different considerations. Re-training RL policies for each reward weight configuration to analyze its sensitivity is computationally expensive. Instead, here we examine how reward outcomes change when modifying reward weights after training a policy with an initial weight of $w$ = 0.1 (Figure~\ref{fig:w_sens}). 

\begin{figure*} [htbp]
    \centering
    \includegraphics[width=0.7\textwidth]{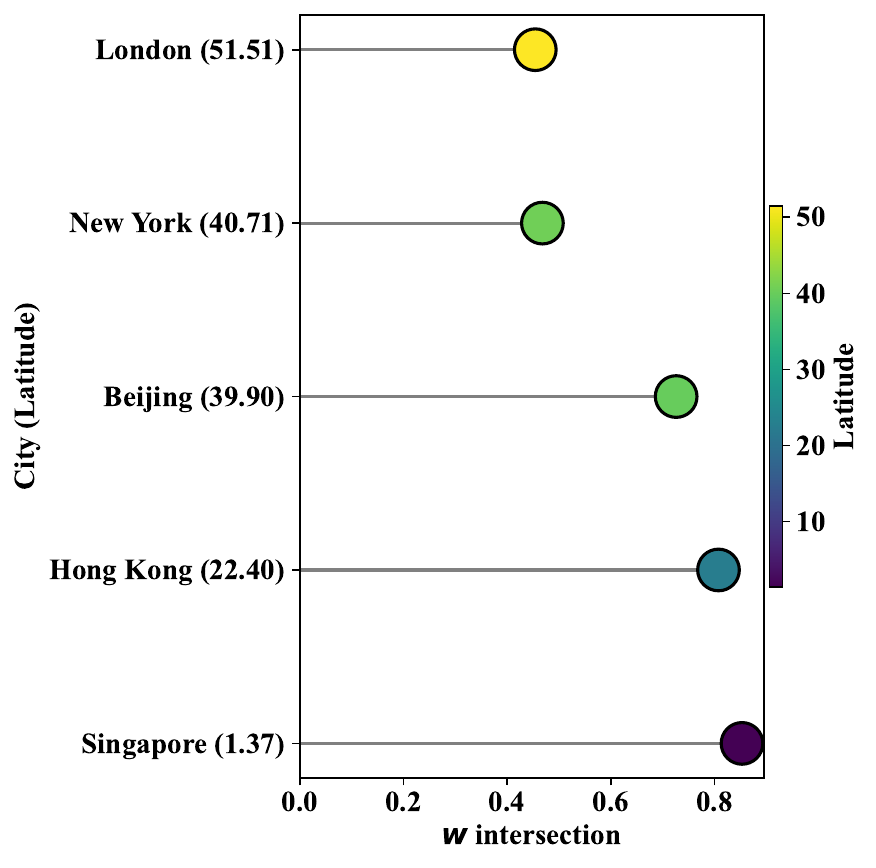}
    \caption{Weight intersection vs. latitude.}
    \label{fig:w_lat}
\end{figure*}

With increasing $w$, the reward of the default strategy rises rapidly, eventually surpassing the reward achieved by RL strategies when $w$ reaches a certain threshold. As $w$ increased, most rewards (both for RL and baselines) also increased, except for the RL strategies in New York. The critical thresholds (the intersection points between the RL and default strategies) were identified, in which the default strategy becomes more advantageous than the RL strategy when $w$ exceeds the threshold. For London and New York, the intersection occurred around 0.5, meaning the RL strategies remain advantageous for approximately half of the weight range. For Beijing, Hong Kong, and Singapore, the intersection appears at a higher weight. Interestingly, the value of the intersection point seems related to latitude (a key factor related to background climate) (Figure~\ref{fig:w_lat}). In higher-latitude regions, the intersection occurs at a lower weight value, suggesting that the advantage of a strategy trained with a 0.1 weight may be less pronounced in higher-latitude cities compared to those at lower latitudes. 

The reward formula indicates that when $w$ = 0, the reward corresponds to the thermal comfort component, and when $w$ = 1, the reward is the energy consumption component. Figure~\ref{fig:w_sens} shows that the RL strategy provides a trade-off---increasing the comfort component at the cost of higher energy consumption. For the default strategy, it places greater emphasis on the energy term. In London and New York, the difference of the comfort component in RL strategy and default strategy is roughly equal to the difference of energy component, indicating that the comfort gains from RL come at a nearly equivalent increase in energy consumption. This balance explains why the intersection point of $w$ is around 0.5. However, in the other three cities, the comfort gains from RL significantly outweigh the penalty from increased energy consumption. 

Although we have conducted an in-depth study on the weights of the RL reward function under a fixed training weight of $w$ = 0.1, there remains room for improving the reward function design. Ideally, RL policies should be trained under different weight settings before comparison. Additionally, RL training in this study employed a linear comfort function based on temperature, yet the relationship between comfort and temperature may not be inherently linear. Future studies could integrate physiological indicators to refine the reward function. 

\subsection{Policy transfer across cities}

\begin{figure*} [htbp]
    \centering
    \includegraphics[width=\textwidth]{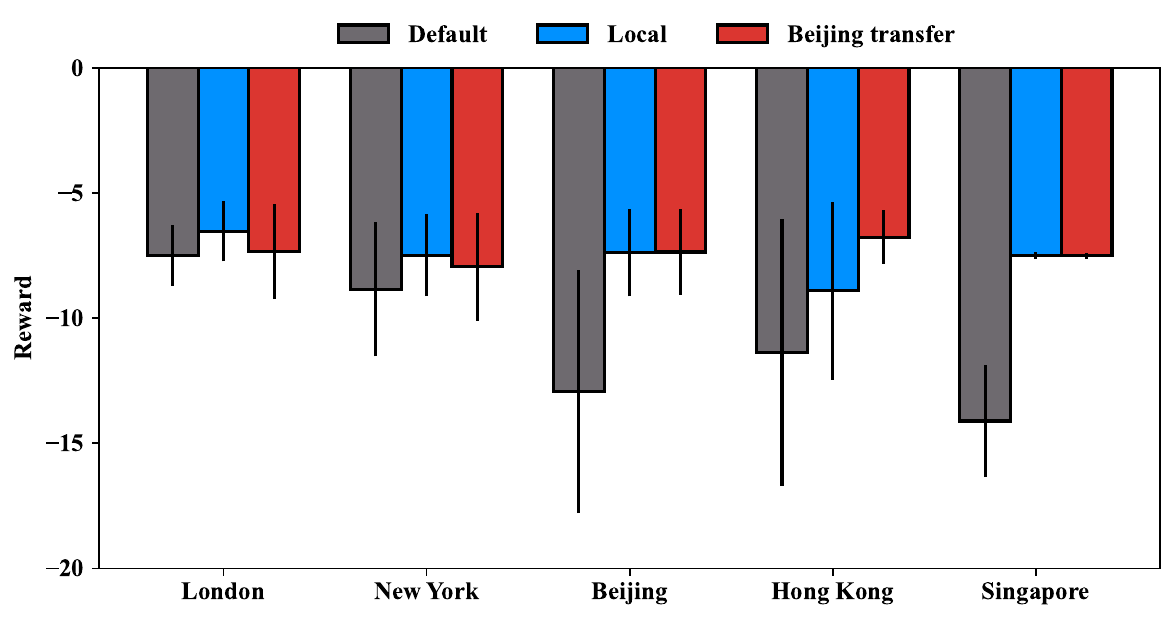}
    \caption{Transferability of the model trained in Beijing. Default is the simulation ran with default setting of CLMU. Local is the simulation ran with the RL model trained in the same city. Beijing transfer indicates the simulation ran with RL model trained in Beijing. }
    \label{fig:trans_beijing}
\end{figure*}

Evaluating the transferability of the model provides insights into the scalability of RL strategies across different cities. When a model exhibits strong transferability, it can be directly applied to new cities without building the model from scratch. Given Beijing’s wide temperature range---cold winters and hot summers---it was selected to evaluate the model's transferability. The results show that the RL model trained in Beijing successfully transferred to other cities, outperforming the default strategy with higher rewards (Figure~\ref{fig:trans_beijing}). When transferred to colder cities---New York and London---the model's performance was slightly lower than the local models, while in warmer cities---Hong Kong and Singapore, its performance was similar or improved (Figure~\ref{fig:trans_beijing}). This variability highlights the importance of background climate for the transferability of the RL model. 

The transferability across all five cities was further evaluated. Results show that most local models achieved higher rewards than the baselines when transferred, except for the RL model trained in Singapore (Figure~S9). Singapore is a typical hot climate city with stable weather for the whole year, contrasting with the other four cities. This likely limits the transferability of the Singapore RL model to other climates.

\begin{figure} [htbp]
    \centering
    \includegraphics[width=0.7\columnwidth]{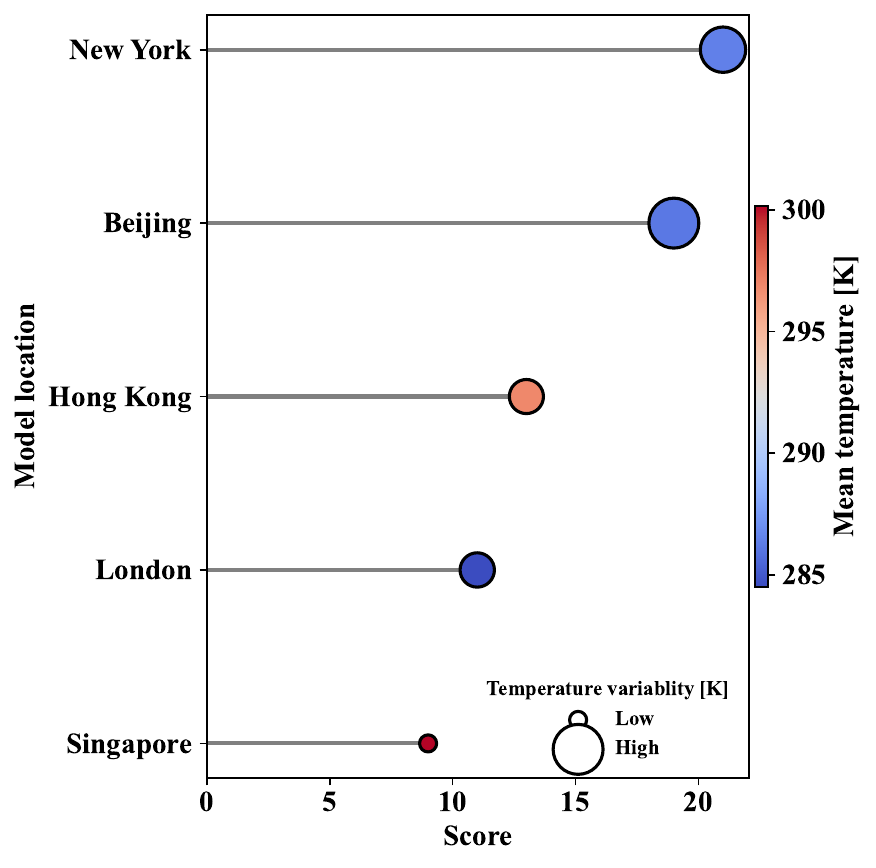}
    \caption{Transferability score of models. The color indicates the mean atmospheric forcing temperature. Size represents the variability (one standard error) of the atmospheric temperature. }
    \label{fig:trans_rank}
\end{figure}

A transferability score was assigned by summing the performance scores of each model across all five cities, with higher scores indicating better overall transferability. Specifically, for each city, the model's performance is evaluated based on its application across different cities, with scores ranging from 1 to 5, where a higher score indicates better performance (e.g., a score of 5 represents the best performance, while a score of 1 represents the worst). These individual scores across all five cities are then summed up for each model. This generates an overall transferability score for each model based on its performance across the five cities, and a higher score indicates better overall transferability. The overall score from best to worst transferability was New York $>$ Beijing $>$ Hong Kong $>$ London $>$ Singapore (Figure~\ref{fig:trans_rank}). Models trained in cities with a higher temperature variability, such as New York and Beijing, may transfer better (Figure~\ref{fig:trans_rank}). Therefore, by identifying an appropriate region for RL model training, it is possible to extend its applicability to other cities and get better results. This result supports that city-to-city learning may be important in applying the climate-sensitive technology (RL-based HVAC control in this study). 

\newpage
\section{Conclusions}

In this study, an urban climate model with a BEM was employed to evaluate how RL-based HVAC control affects indoor climate and local urban climate in five cities. By using RL for HVAC control, thermal comfort can be enhanced while maintaining energy efficiency. These improvements gained from RL are closely linked to the background climate and exhibit seasonal variations for some cities. This is critical for cities aiming to implement RL-based HVAC control, as it emphasizes the needs to account for background climate when deploying such technology. 

The responses of indoor climate and local urban climate to RL strategies varied notably across cities and correlated with background climates too. These findings are consistent with other studies demonstrating that the urban heat island effect and the effectiveness of urban adaptation strategies are strongly influenced by the background climate \citep{MANOLIETAL19Nature, WANGETAL20Geophys.Res.Lett.b, ZHAOETAL14Natureb, ZHAOETAL23Environ.Res.Lett.}. Consequently, policymakers and stakeholders should consider developing city-specific RL strategies for HVAC control. Meanwhile, cities with larger seasonal temperature variations tend to exhibit greater transferability, meaning that RL-based control strategies trained in such cities may be more adaptable to other locations. This suggests that while RL strategies should incorporate city-specific considerations, training in a climate-diverse city could enhance their scalability across different regions.

While this study demonstrates the potential of RL-based HVAC control strategies within the urban climate modeling framework, several limitations remain. First, the simplified BEM used in this study \citep{OLESONFEDDEMA20J.Adv.Model.EarthSyst.b, LIETAL24J.Adv.Model.EarthSyst.a} omits details such as building floors, room counts, and window configurations, potentially introducing simulation bias. Future work should aim to enhance BEM simulation capabilities within climate models, allowing for a more detailed representation of indoor and outdoor climate interactions. Second, the weight in this study was predefined preferences, a common approach for solving multi-objective RL problems \citep{XUETAL20Proc.37thInt.Conf.Mach.Learn.a}. However, fixed weights are susceptible to personal preferences and limited knowledge. Future work could employ multi-objective RL to reduce this dependency \citep{FELTENETAL23Proc.37thInt.Conf.NeuralInf.Process.Syst., XUETAL20Proc.37thInt.Conf.Mach.Learn.a}, reframing the reward function as a multi-objective optimization problem to identify a Pareto-optimal solution under varying weight scenarios. Lastly, the global generalizability of our findings remains uncertain, as this study focused on five cities and used an offline climate model, preventing urban land surface-induced climate feedback to the atmosphere \citep{SUNETAL24J.Adv.Model.EarthSyst.}. Such feedback could significantly influence future climate conditions. As HVAC control technology scales, it will become an important subprocess in urban biophysics, potentially impacting climate at larger scales. Future research should assess RL strategies globally by leveraging fully coupled simulations to evaluate their climate impacts and Earth system impacts, similar to studies on solar panel installations \citep{HUETAL16NatureClimateChange}.

\section{Acknowledgments}
This work was supported by The Royal Society \url{IEC\NSFC\223167} - International Exchanges 2022 Cost Share (NSFC).

\section{Code and data availability}
Code and data are available at \url{https://github.com/envdes/code_CLMU_HVAC_RL}.

\newpage
\bibliographystyle{elsarticle-harv} 
\bibliography{paper_clmurl}

\newpage
\begin{appendices}
\section*{Supporting Information}

\journal{arXiv}

\setcounter{section}{0}
\setcounter{figure}{0}
\setcounter{table}{0}
\setcounter{equation}{0}

\renewcommand{\thesection}{S\arabic{section}}
\renewcommand{\thefigure}{S\arabic{figure}}
\renewcommand{\thetable}{S\arabic{table}}
\renewcommand{\theequation}{S\arabic{equation}}

\renewcommand{\thefigure}{S\arabic{figure}}
\renewcommand{\thetable}{S\arabic{table}}
\renewcommand{\theequation}{S\arabic{equation}}

\section{Details of RL agents}

\subsection{Q-learning}
Q-learning learning updates the Q function as follow:

\begin{equation}
Q(s, a) \leftarrow Q(s, a) + \alpha \left[ r + \gamma \max_{a'} Q(s', a') - Q(s, a) \right]
\end{equation}

\noindent
where $Q(s,a)$ is the current Q-value for taking action $a$ in state $s$. $\alpha$ is the learning rate, controlling the amplitude of updates. $r$ is the reward received after taking action $a$. $\gamma$ is the discount factor that determines the importance of future rewards. $s'$ The next state resulting from the selected action. $\max_{a'} Q(s', a')$ is the operator to get the maximum Q-value over all possible actions in the next state. This equation iteratively updates the Q-value through the interaction with the environment. 

Q-leaning algorithm starts with the initialization of the Q-function (a table). We used zero values to initialize the $Q(s, a)$. In this study, the reward is always negative, and the zero initialization will make all the possible discretized actions ($a$) get explored. We implemented an exponentially decaying rate to ensure ample exploration with a maximum exploration rate of 1.0, a minimum exploration rate of 0.01, and a decay rate of 0.01. The discount factor $\gamma$ is set to 0.99. 

\begin{algorithm} \label{algo:qlearning}
\caption{Q-learning Algorithm}
\begin{algorithmic}[1]
\Require State space $\mathcal{S}$, action space $\mathcal{A}$, learning rate $\alpha$, discount factor $\gamma$, initial $\epsilon = \text{max\_epsilon}$, minimum $\epsilon = \text{min\_epsilon}$, decay rate $\lambda$, total epochs $E$
\Ensure Optimized Q-function $Q(s, a)$
\State Initialize $Q(s, a)$ with zero values for all $s \in \mathcal{S}, a \in \mathcal{A}$
\For{epoch $e = 1$ to $E$}
    \State Observe initial state $s$
    \State Compute $\epsilon$ for this epoch:
    \[
    \epsilon = \text{min\_epsilon} + (\text{max\_epsilon} - \text{min\_epsilon}) \cdot \exp(-\lambda \cdot e)
    \]
    \While{not terminal state}
        \State Select action $a$ using $\epsilon$-greedy policy
        \State Execute action $a$, observe reward $r$ and next state $s'$
        \State Update Q-value:
        \[
        Q(s, a) \gets Q(s, a) + \alpha \left[ r + \gamma \max_{a'} Q(s', a') - Q(s, a) \right]
        \]
        \State Set $s \gets s'$
    \EndWhile
\EndFor
\end{algorithmic}
\end{algorithm}

\subsection{DQN}

DQN learns the Q-function using neural networks (parameterized with $\theta$). DQN trains the target network and the master network ($\theta$) simultaneously, but the target network ($\theta'$) is updated periodically. The procedure of DQN is shown in algorithm~\ref{algo_dqn}. After initialization, the algorithm will collect transitions of state ($s$), action ($a$), reward ($r$), and the next state ($s'$), in which $s$, ($r$), and $s'$ is from the environment, and $a$ is from the network with $\theta$. These transitions will be stored in a replay buffer ($\mathcal{D}$) with size $B$. The experiences from $\mathcal{D}$ are used to calculate the temporal difference (TD) target ($y_i$),

\begin{equation}
    y_i = r_i + \gamma \max\limits_{a'} Q_{\theta'}(s_i', a')
\end{equation}

\noindent
where $\gamma$ is the discount factor. Notably, when $s'$ is terminal, $y_i = r_i$. $y_i$ is used to calculate the loss of $Q_{\theta}$, which is represented by mean squared value of $y_i$ and $Q_{\theta}$. This loss is then applied to update $\theta$ using the descent gradient method. The $Q_{\theta'}$ is then updated using the $\theta$ with a soft update factor $\tau$. 

The neuron networks are multilayer perceptrons. The observation space of the environment determines the input dimension (5 in this study) of the network. The features are mapped to a 128-dimensional hidden layer through the first linear layer, followed by a rectified linear unit (ReLU) activation. Then, the second linear layer, also with a dimension of 128, is applied with another ReLU activation. Finally, the features are mapped to the action space dimension (3 in this study) through the output layer. The parameters of DQN in this study are $2.5\times 10^{-4}$ of learning rate for updating $\theta$ using the descent gradient method, 10000 of $B$, 100 of $M$, 1.0 of $\tau$, 0.99 of $\gamma$, and 10000 of $n$. Technically, the $\theta$ is updated when the environment is performed in every 10 steps. The exploration rate is updated using the linear decay method. To compare among different cities, these parameters remain the same. These network architectures and parameters are default as CleanRL repository (\url{https://github.com/vwxyzjn/cleanrl}) \citep{HUANGETAL22J.Mach.Learn.Res.}

\begin{algorithm}[H]
\caption{DQN Algorithm}\label{algo_dqn}
\begin{algorithmic}[2]
\Require State space $\mathcal{S}$, action space $\mathcal{A}$, learning rate $\alpha$, discount factor $\gamma$, replay buffer size $B$, batch size $M$, total steps $TS$, learning start step $n$, soft update factor $\tau$
\Ensure Trained Q-network $Q_\theta(s, a)$
\State Initialize Q-network with parameters $\theta$
\State Initialize target Q-network $Q_{\theta'}(s, a)$ with parameters $\theta^{'}= \theta$
\State Initialize replay buffer $\mathcal{D}$ to capacity $B$

\For{step $t = 1$ to $TS$}
    \State Observe initial state $s$
    \State Compute exploration rate $\epsilon$ 
    \If{not terminal state}
        \State Select action $a$ using $\epsilon$-greedy policy with $\epsilon$
        \State Execute action $a$, observe reward $r$ and next state $s'$
        \State Store transition $(s, a, r, s')$ to $\mathcal{D}$
        \If{$t \geq n$}
        \State Sample a random mini-batch of $M$ transitions from $\mathcal{D}$
        \State Compute target Q-value ($y_i$) for each sample
        \State Update Q-network by minimizing loss:
        \[
        L(\theta) = \frac{1}{M} \sum_{i=1}^M \left( y_i - Q_\theta(s_i, a_i) \right)^2
        \]
        \State Update $\theta$ using descent gradient ($\nabla_\theta L(\theta)$)
        \State Set $s \gets s'$
        
        \EndIf
        \State Update target network parameters periodically: $\theta^{'} \gets (1-\tau)\theta$
    \EndIf
\EndFor
\end{algorithmic}
\end{algorithm}

\subsection{SAC}

SAC trains the Q-network ($Q_{\theta}(s, a)$) and policy network $\pi_\phi(a|s)$, in which action $a$ is predicted from $\pi_\phi(a|s)$ and the value of $a$ is judged by $Q_{\theta}(s, a)$. This framework is called actor-critic, where the actor is $\pi_\phi(a|s)$ and the critic is $Q_{\theta}(s, a)$. In application, only $\pi_\phi(a|s)$ is used. During training, $Q_{\theta}(s, a)$ is trained using the same way as DQN but with a different target Q-value ($y_i$), which is given as follows:

\begin{equation}\label{sac_td}
    y_i = r_i + \gamma \min_{j=1,2} Q_{\theta_j'}(s_i', a_i') - \alpha \log \pi_\phi(a_i'|s_i')
\end{equation}
\noindent
where $Q_{\theta_j'}$ is the target Q-network but with different parameters. There are two Q-networks trained and the minor is used to calculate $y_i$ to lower the overestimation bias. $\log \pi_\phi(a_i'|s_i')$ is the policy entropy, which is used to control the exploration and exploitation. The magnitude of policy entropy is control by temperature parameter ($\alpha$). The two Q-network $Q_{\theta_j}$ are then updated by

\begin{equation}\label{sac_Lq}
    L_Q(\theta_j) = \frac{1}{M} \sum_{i=1}^M \left( y_i - Q_{\theta_j}(s_i, a_i) \right)^2
\end{equation}

\noindent
where $L_Q(\theta_j)$ is the loss of Q-network. These Q-networks are updated simultaneously by combining two losses. The two target Q-network ($Q_{\theta_j'}$) are updated using $\theta_j$ with $\tau$ periodically, respectively. 

The policy network is updated by

\begin{equation}
    J_\pi(\phi) = \frac{1}{M} \sum_{i=1}^M \left( \alpha \log \pi_\phi(a_i|s_i) - min_{j=1,2}Q_{\theta_j}(s_i, a_i) \right).
\end{equation}

\noindent
This objective maximizes the values of policy and ensures entropy. 

The $\alpha$ is also updated using 

\begin{equation}
    L_\alpha = -\alpha \mathbb{E}_{a \sim \pi_\phi} \left[ \log \pi_\phi(a|s) + \mathcal{H}_\text{target} \right]
\end{equation}

\noindent
where $\mathcal{H}_\text{target}$ is the target entropy, calculated based on the action space. 

The neuron networks of Q-networks are also multilayer perceptrons, which have the same architecture as DQN, but the hidden layer dimensions are 256. The actor network is a Gaussian policy network, which outputs the mean and log standard deviation of policy. The architecture is the same as Q-networks but has two output heads (each is one-dimensional). The Q-networks and policy network learning rate are $10^{-3}$ and $3\times10^{-4}$, respectively. Note that the policy network is updated every two steps to stabilize training by reducing the frequency of updates and allowing the Q-value network to learn more effectively. The target Q-network frequency is set to 1, updating the target network every step, promoting more consistent updates. The discount factor $\gamma$ is 0.99, the replay buffer size $B$ is $10^6$, the batch size $m$ is set to 256, the learning start step $n$ is set to 5000, soft update factor $\tau$ is 0.005, and the temperature parameter $\alpha$ is set to 0.2. The network architecture and parameters keep the same as CleanRL repository (\url{https://github.com/vwxyzjn/cleanrl}) \citep{HUANGETAL22J.Mach.Learn.Res.}.

\begin{algorithm}[H]
\caption{SAC Algorithm}\label{algo_sac}
\begin{algorithmic}[2]
\Require State space $\mathcal{S}$, action space $\mathcal{A}$, learning rates $\alpha_\pi$, $\alpha_Q$, $\alpha_\text{entropy}$, discount factor $\gamma$, replay buffer size $B$, batch size $M$, temperature parameter $\alpha$, soft update factor $\tau$, total step $TS$, learning start step $n$
\Ensure Trained policy network $\pi_\phi(a|s)$ and Q-networks $Q_{\theta_1}(s, a)$, $Q_{\theta_2}(s, a)$
\State Initialize policy network $\pi_\phi(a|s)$ with parameters $\phi$
\State Initialize Q-networks $Q_{\theta_1}(s, a)$, $Q_{\theta_2}(s, a)$ with parameters $\theta_1$, $\theta_2$
\State Initialize target Q-networks $\theta_1' \gets \theta_1$, $\theta_2' \gets \theta_2$
\State Initialize replay buffer $\mathcal{D}$ to capacity $B$

\For{step $t = 1$ to $TS$}
    \State Observe initial state $s$
    \If{not terminal state}
        \State Select action $a \sim \pi_\phi(a|s)$
        \State Execute action $a$, observe reward $r$ and next state $s'$
        \State Store transition $(s, a, r, s')$ to $\mathcal{D}$
        \If{$t \geq n$}
            \State Sample a mini-batch of $M$ transitions from $\mathcal{D}$
            
            \State Compute target Q-value:
            \[
            y_i = r_i + \gamma \min_{j=1,2} Q_{\theta_j'}(s_i', a_i') - \alpha \log \pi_\phi(a_i'|s_i')
            \]
            
            \State Update Q-networks by the loss for $j=1,2$:
            \[
            L_Q(\theta_j) = \frac{1}{M} \sum_{i=1}^M \left( y_i - Q_{\theta_j}(s_i, a_i) \right)^2
            \]

            \State Update policy network by:
            \[
            J_\pi(\phi) = \frac{1}{M} \sum_{i=1}^M \left( \alpha \log \pi_\phi(a_i|s_i) - min_{j=1,2}Q_{\theta_j}(s_i, a_i) \right)
            \]

            \State Adjust temperature parameter $\alpha$ :
            \[
            L_\alpha = -\alpha \mathbb{E}_{a \sim \pi_\phi} \left[ \log \pi_\phi(a|s) + \mathcal{H}_\text{target} \right]
            \]
            \State Update $\alpha$ via gradient descent on $L_\alpha$
            
            \State Update target Q-network parameters periodically:
            \[
            \theta_j' \gets \tau \theta_j + (1 - \tau) \theta_j', \quad j=1, 2
            \]
        \EndIf
        \State Set $s \gets s'$
    \EndIf
\EndFor
\end{algorithmic}
\end{algorithm}

\begin{table}[htbp]
\caption{Main inputs for Python BEM}
\footnotesize
\centering
\begin{tabular}{|p{2cm}|p{10cm}|}
\hline
\textbf{Category}               & \textbf{Variable/Parameter}                                                                                                                                                        \\ \hline
\textbf{Building Data}          & \begin{itemize}
                                   \item Building height (m)
                                   \item Urban canyon width-to-height ratio (for estimating building width) (Unitless)
                                   \item Roof area fraction (Unitless)
                                   \item Thickness and thermal conductivity of the innermost roof and wall layers (m)
                                   \item Building ventilation rate (air exchanges per hour)
                                   \item Maximum and minimum internal building air temperature (K)
                                   \item Ventilation rate (m$^3$/s)
                                 \end{itemize}                                                                                                                                       \\ \hline
\textbf{Temperature Forcing}    & \begin{itemize}
                                   \item Temperatures:
                                     \begin{itemize}
                                       \item Urban canopy air temperature
                                       \item Roof temperature at inner node depth (K)
                                       \item Sunlit wall temperature at inner node depth (K)
                                       \item Shade wall temperature at inner node depth (K)
                                     \end{itemize}
                                   \item Previous time step temperatures:
                                     \begin{itemize}
                                       \item Roof inside surface temperature (K)
                                       \item Sunlit wall inside surface temperature (K)
                                       \item Shaded wall inside surface temperature (K)
                                       \item Floor temperature (K)
                                       \item Internal building air temperature (K)
                                       \item Roof temperature at inner node depth (K)
                                       \item Sunlit wall temperature at inner node depth (K)
                                       \item Shade wall temperature at inner node depth (K)
                                     \end{itemize}
                                 \end{itemize}                                                                                                                                       \\ \hline
\end{tabular}
\label{tab:bem_inputs}
\end{table}

\begin{table}[htbp]
\caption{Outputs, and validation details for Python BEM}
\footnotesize
\centering
\begin{tabular}{|p{2cm}|p{10cm}|}
\hline
\textbf{Category}               & \textbf{Variable/Parameter}                                                                                                                                                        \\ \hline
\textbf{Model Outputs}          & \begin{itemize}
                                   \item Updated temperatures at the next time step:
                                     \begin{itemize}
                                       \item Roof inside surface temperature
                                       \item Sunlit wall inside surface temperature
                                       \item Shaded wall inside surface temperature
                                       \item Floor temperature
                                       \item Internal building air temperature
                                     \end{itemize}
                                   \item Flux calculations:
                                     \begin{itemize}
                                       \item Building heat flux (from changes in building air temperature)
                                       \item Urban air conditioning flux
                                       \item Urban heating flux
                                       \item Ventilation flux
                                     \end{itemize}
                                 \end{itemize}                                                                                                                                       \\ \hline
\textbf{Source of Data} & \begin{itemize}
                                       \item Building data is obtained from CLMU simulation surface data.
                                       \item Urban canopy air temperature and previous time step temperatures at inner node depth: provided by complete CLMU simulation
                                       \item Roof, sunlit wall, shaded wall, floor, and internal building air temperatures: obtained through iterative calculations
                                     \end{itemize}                                                                                                                              \\ \hline
\textbf{Validation}             & \begin{itemize}
                                   \item Validation provided in Figure~S2--6.
                                   \item Error in building air temperature calculation: on the order of $10^{-5} \, \text{K}$, likely due to rounding errors in floating-point numbers (differences in algorithms, precision settings, or hardware architectures)
                                 \end{itemize}                                                                                                                                       \\ \hline
\end{tabular}
\label{tab:bem_outputs}
\end{table}

\begin{figure}[htbp]
  \centering
  \includegraphics[width=1\textwidth]{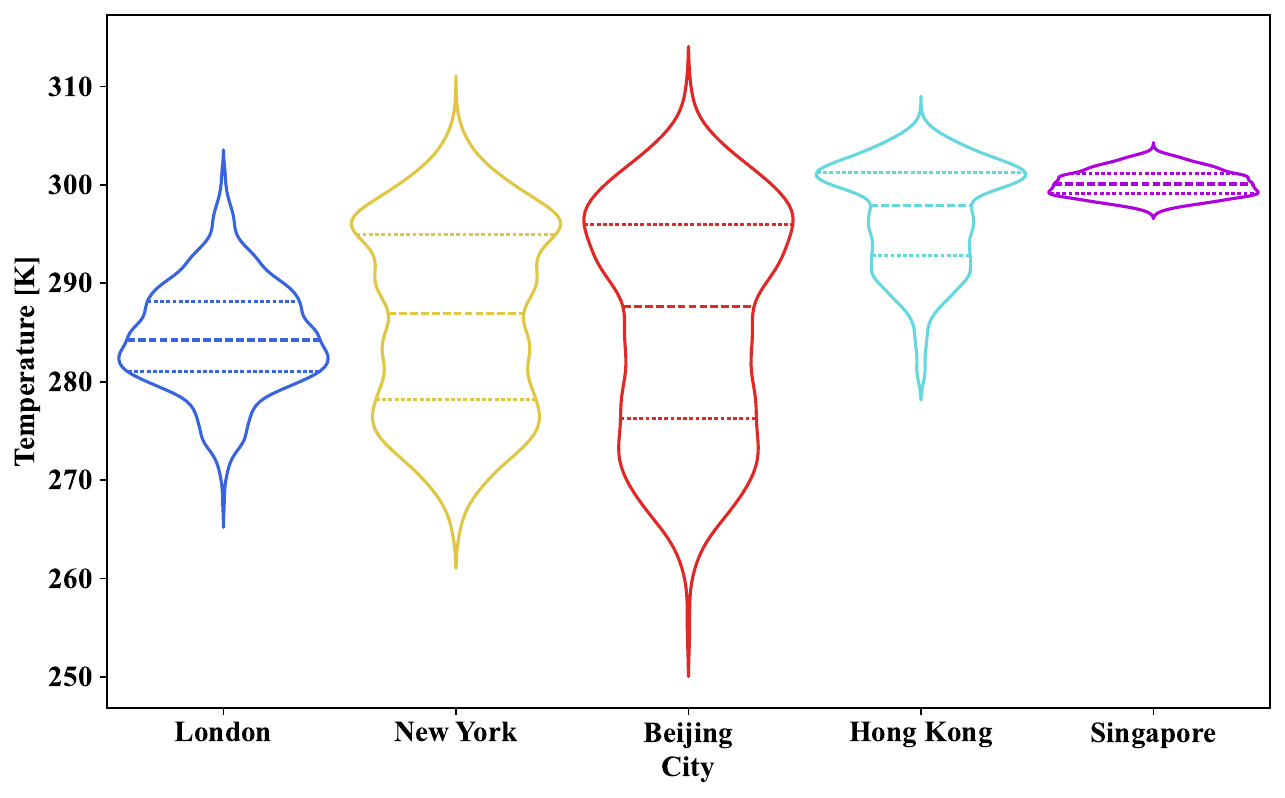}
  \caption{Forcing temperature distribution.}
  \label{figs:temp_f}
\end{figure}

\begin{figure}[htbp]
  \centering
  \includegraphics[width=1\textwidth]{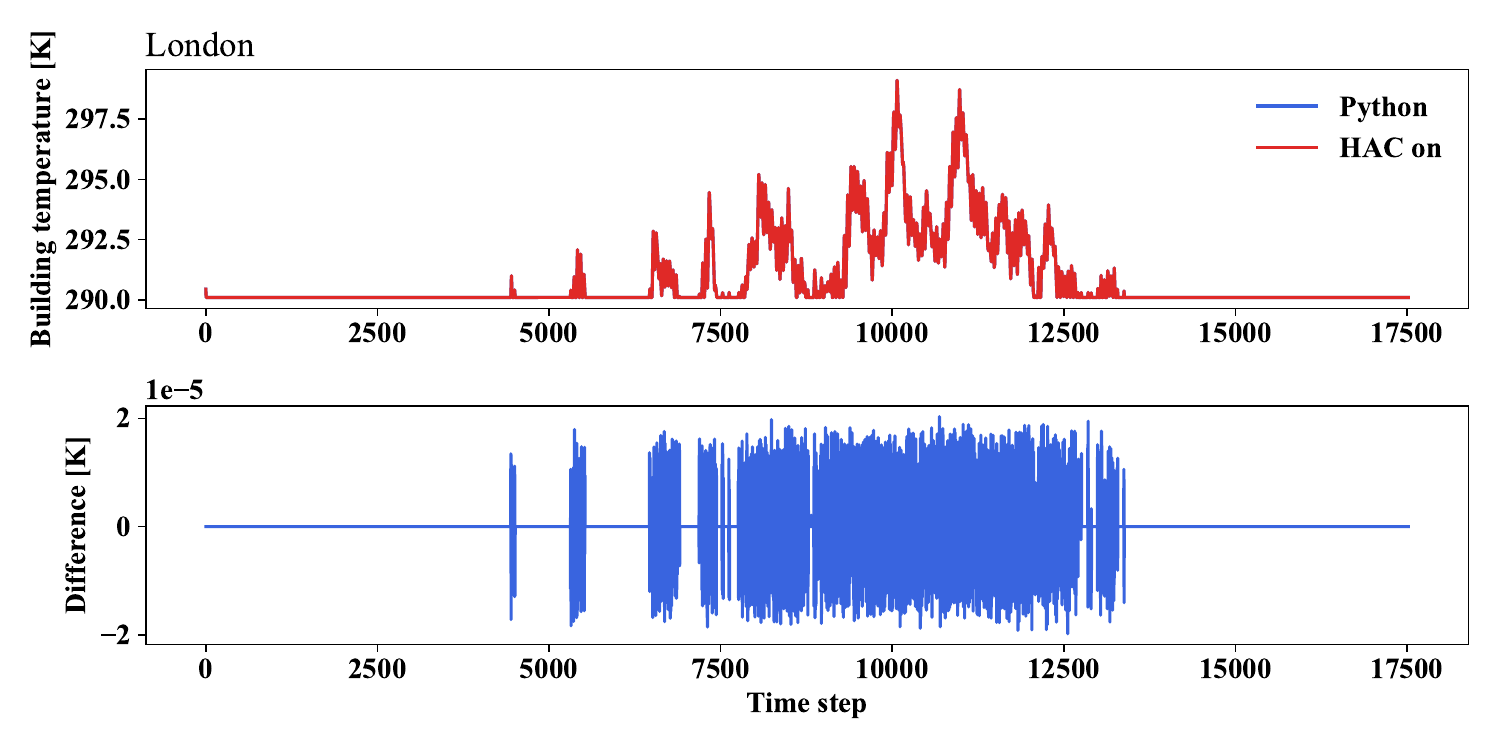}
  \caption{Validation of CLMU/BEM and Python version BEM in London.}
  \label{figs:val-london}
\end{figure} 

\begin{figure}[htbp]
  \centering
  \includegraphics[width=1\textwidth]{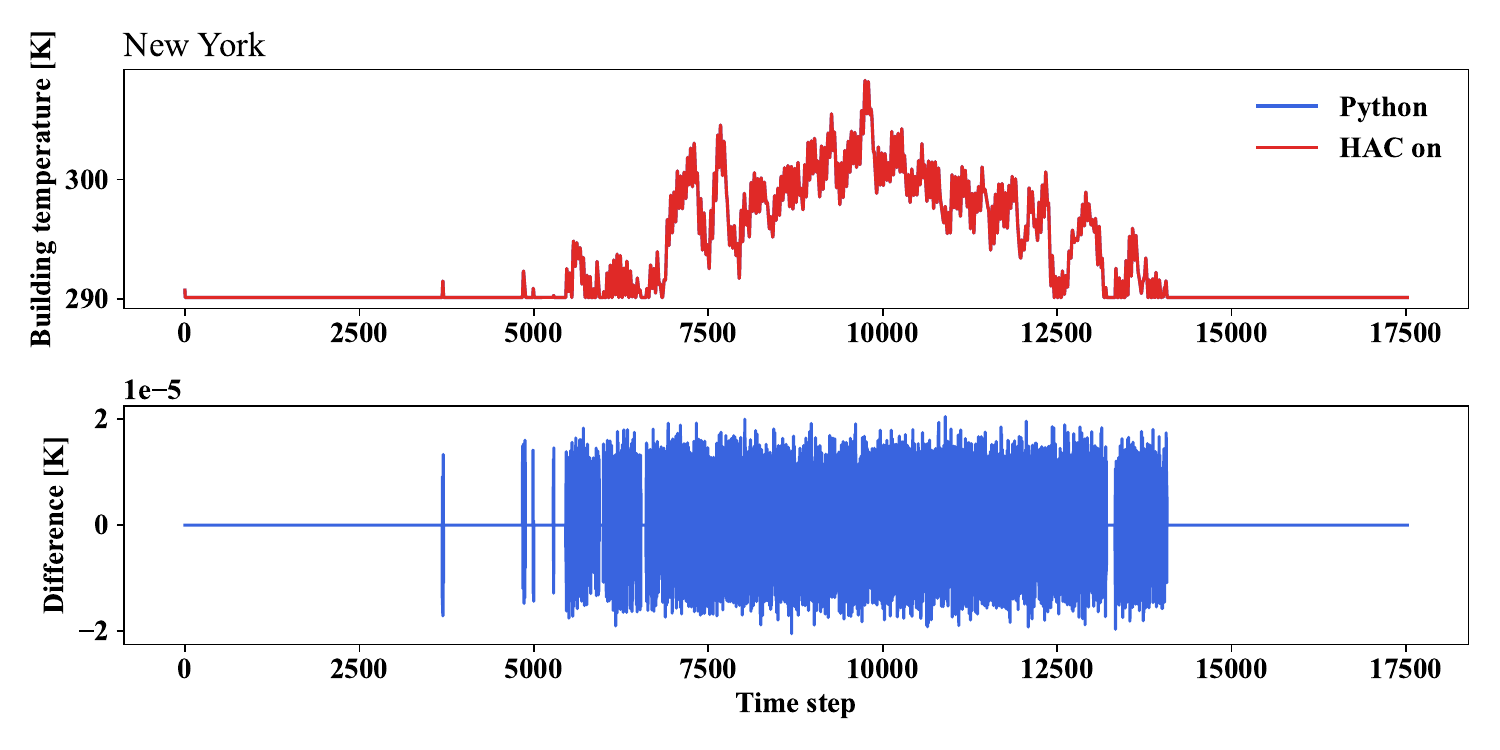}
  \caption{Validation of CLMU/BEM and Python version BEM in New York.}
  \label{figs:val-newyork}
\end{figure}

\begin{figure}[htbp]
  \centering
  \includegraphics[width=1\textwidth]{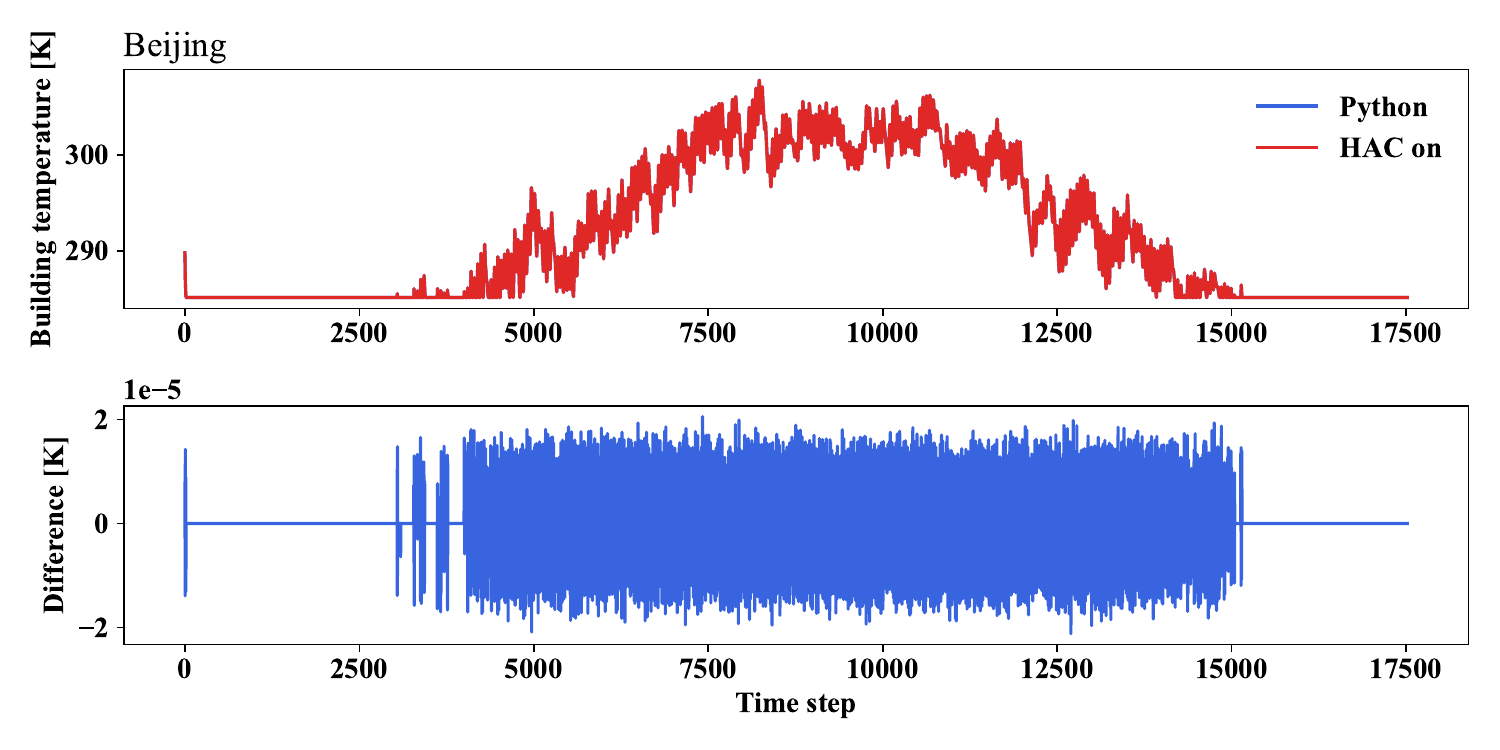}
  \caption{Validation of CLMU/BEM and Python version BEM in Beijing.}
  \label{figs:val-beijing}
\end{figure}

\begin{figure}[htbp]
  \centering
  \includegraphics[width=1\textwidth]{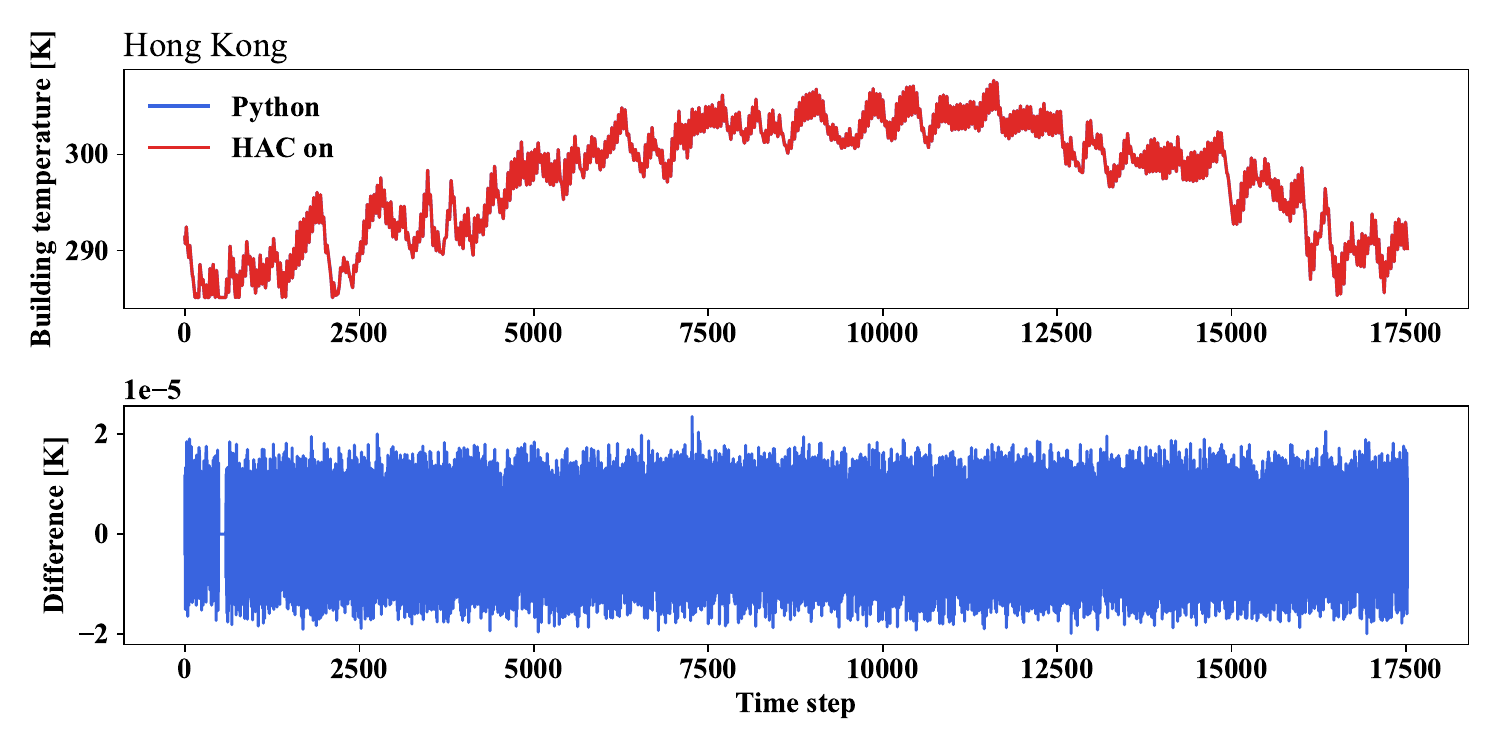}
  \caption{Validation of CLMU/BEM and Python version BEM in Hong Kong.}
  \label{figs:val-hongkong}
\end{figure}

\begin{figure}[htbp]
  \centering
  \includegraphics[width=1\textwidth]{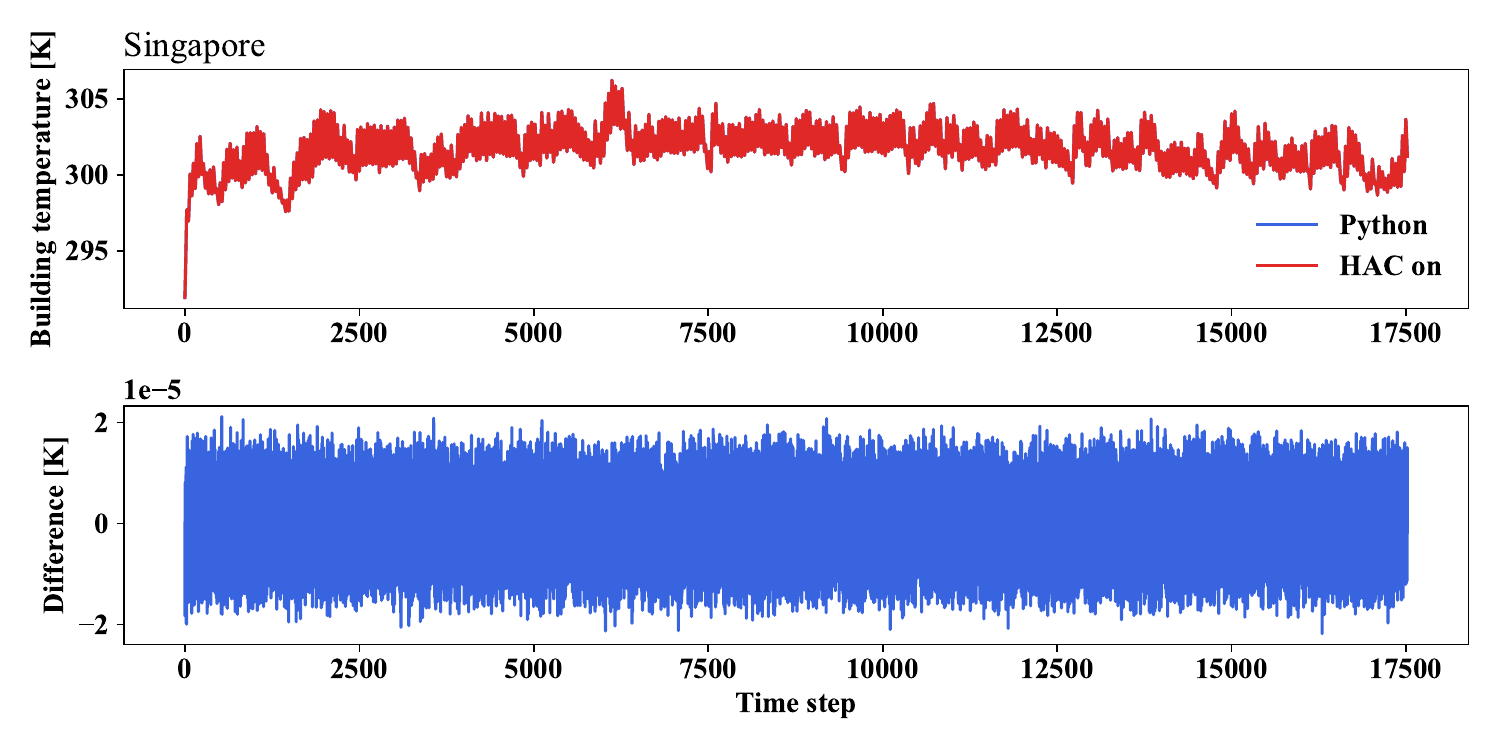}
  \caption{Validation of CLMU/BEM and Python version BEM in Singapore.}
  \label{figs:val-singapore}
\end{figure}

\newpage
\begin{figure}[htbp]
  \centering
  \includegraphics[width=1\textwidth]{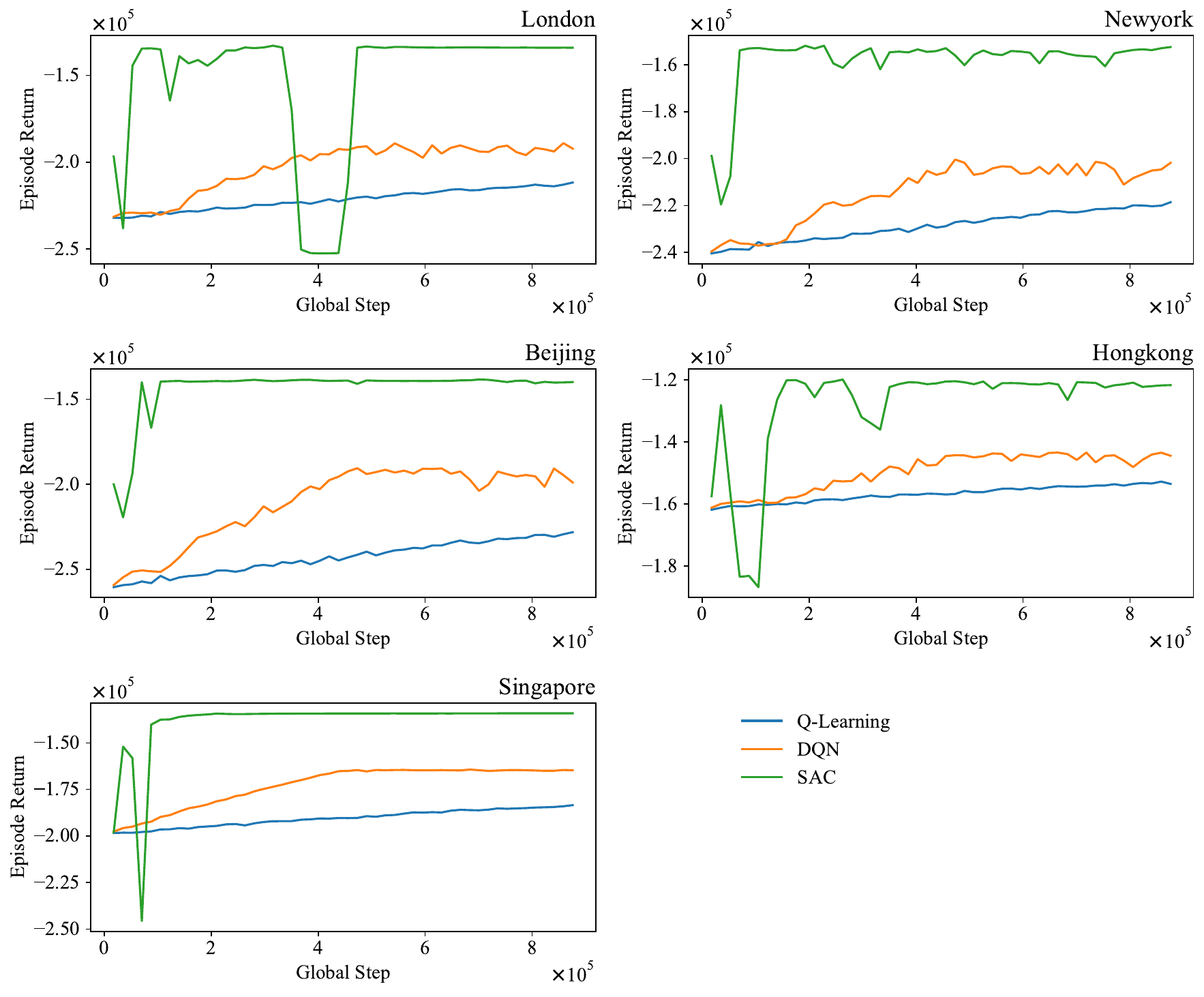}
  \caption{RL model training processes.}
  \label{figs:training}
\end{figure}

\begin{figure}[htbp]
  \centering
  \includegraphics[width=1\textwidth]{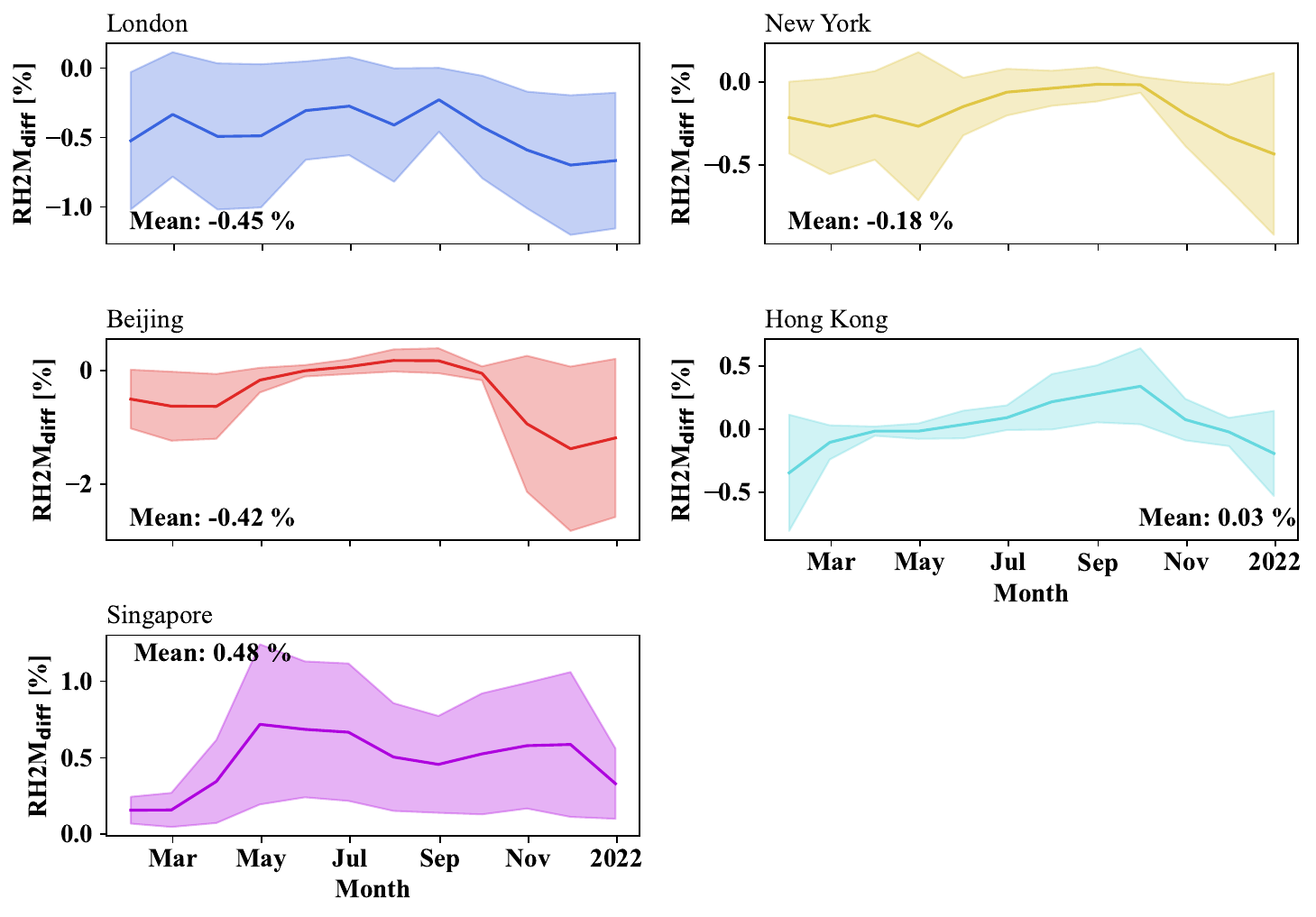}
  \caption{Monthly profile of relative humidity in urban canopy difference in RL cases and CLMU default cases across cities.}
  \label{figs:rh}
\end{figure}


\begin{figure}[htbp]
  \centering
  \includegraphics[width=1\textwidth]{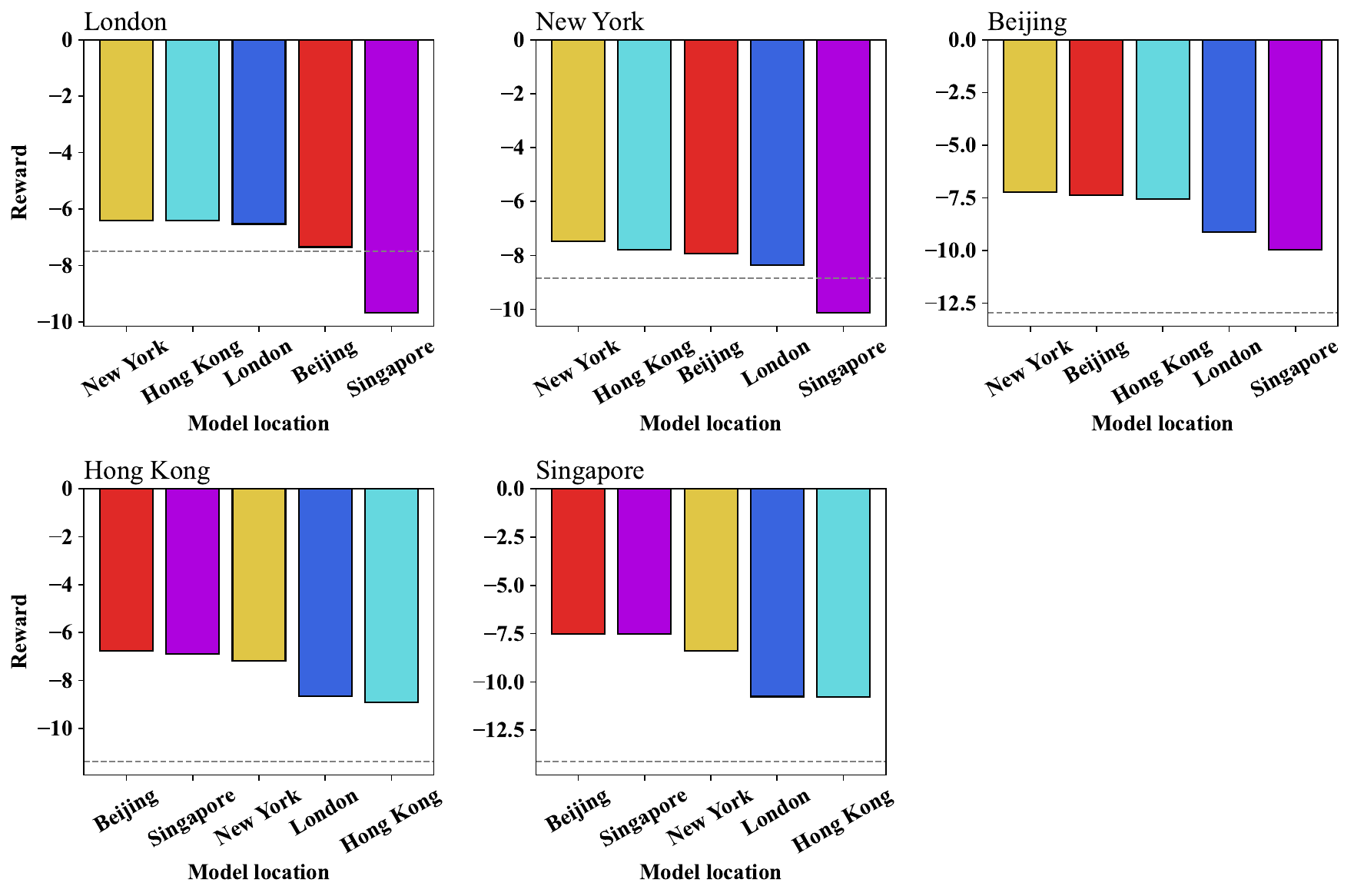}
  \caption{Transferability of models. The gray dash lines are the reward from the default cases.}
  \label{figs:transfer}
\end{figure}

\end{appendices}

\end{document}